\documentclass[journal]{IEEEtran}
\ifCLASSINFOpdf
   \usepackage[pdftex]{graphicx}
\else
\fi
\usepackage{amsmath}
\usepackage{amssymb}
\usepackage{mathtools}
\usepackage{soul}
\usepackage{pifont}
\usepackage{color}

\newcommand{\PLH}{{\mkern-2mu\times\mkern-2mu}}
\usepackage{algorithm}
\usepackage{algorithmic}

\usepackage{booktabs}
\usepackage{multicol,multirow}
\usepackage{makecell}
\usepackage{threeparttable}
\usepackage{tablefootnote}
\usepackage{url}

\usepackage{pifont}
\newcommand{\cmark}{\ding{51}}%
\begin{document}
%
\title{SmartDeal: Re-Modeling Deep Network Weights for Efficient Inference and Training}
%
%
%
\author{Xiaohan Chen$^*$, Yang Zhao$^*$, Yue Wang, Pengfei Xu, Haoran You, Chaojian Li, Yonggan Fu, \\ Yingyan Lin, and Zhangyang Wang
\thanks{X. Chen and Z. Wang are with the Department
of Electrical and Computer Engineering, The University of Texas at Austin, Austin, TX 78712 USA. Email: \{xiaohan.chen, atlaswang\}@utexas.edu.}
\thanks{Y. Zhao, Y. Wang, P. Xu, H. You, C. Li, Y. Fu, and Y. Lin are with Rice University, Houston, TX 77005 USA. Email: \{zy34, yw68, px5, hy34, cl114, yf22, yingyan.lin\}@rice.edu.}
\thanks{The first two authors X. Chen and Y. Zhao contributed equally to this work.}}

%
%

\markboth{Journal of \LaTeX\ Class Files,~Vol.~14, No.~8, August~2015}%
{Shell \MakeLowercase{\textit{et al.}}: Bare Demo of IEEEtran.cls for IEEE Journals}
%



\maketitle

\begin{abstract}

The record-breaking performance of deep neural networks (DNNs) comes with heavy parameter budgets, which leads to external dynamic random-access memory (DRAM) for storage. The prohibitive energy of DRAM accesses makes it non-trivial for DNN deployment on resource-constrained devices, calling for minimizing the movements of weights and data in order to improve the energy efficiency. Driven by this critical bottleneck, we present \textit{SmartDeal}, a hardware-friendly algorithm framework to trade higher-cost memory storage/access for lower-cost computation, in order to aggressively boost the storage and energy efficiency, for both DNN inference \text{and} training. 

The core technique of \textit{SmartDeal} is a novel DNN weight matrix decomposition framework with respective structural constraints on each matrix factor, carefully crafted to unleash the hardware-aware efficiency potential. Specifically, we decompose each weight tensor as the product of a small basis matrix and a large structurally sparse coefficient matrix whose non-zero elements are readily quantized to power-of-2.
The resulting sparse and readily-quantized DNNs enjoy greatly reduced energy consumption in data movement as well as weight storage, while incurring minimal overhead to recover the original weights thanks to the required sparse bit-operations and cost favorable computations. Beyond inference, we take another leap to embrace energy-efficient \textbf{training}, by introducing several customized techniques to address the unique roadblocks arising in training while preserving the \textit{SmartDeal} structures. We also design a dedicated hardware accelerator to fully utilize the new weight structure to improve the real energy efficiency and latency performance. 

We conduct experiments on both vision and language tasks, with nine models, four datasets, and three settings (inference-only, adaptation, and fine-tuning). Our extensive results show that: 1) \underline{being applied to inference}, \textit{SmartDeal} achieves up to 2.44$\times$ improvement in energy efficiency as evaluated via real hardware implementations; 
2) \underline{being applied to training}, \textit{SmartDeal} can lead to 10.56$\times$ and 4.48$\times$ reduction in the storage and the training energy cost, respectively, with usually negligible accuracy loss, compared to state-of-the-art training baselines. Our source codes are available at: \textcolor{blue}{\url{https://github.com/VITA-Group/SmartDeal}}.
\end{abstract}

\begin{IEEEkeywords}
Efficient Machine Learning, Deep Network Training, Data Movement, Hardware Accelerator
\end{IEEEkeywords}

%
\IEEEpeerreviewmaketitle

\section{Introduction}
\label{sec:introduction}
%
%
%
%

\subsection{Background and Core Idea}

\IEEEPARstart{T}{he} performance breakthrough of deep neural networks (DNNs) motivates a growing demand to bring DNNs into storage- and energy-constrained edge devices, such as mobile phones, wearables, and IoT sensors, using domain-specific accelerators. 
\begin{figure}[!t]
\centering
\includegraphics[width=0.96\columnwidth]{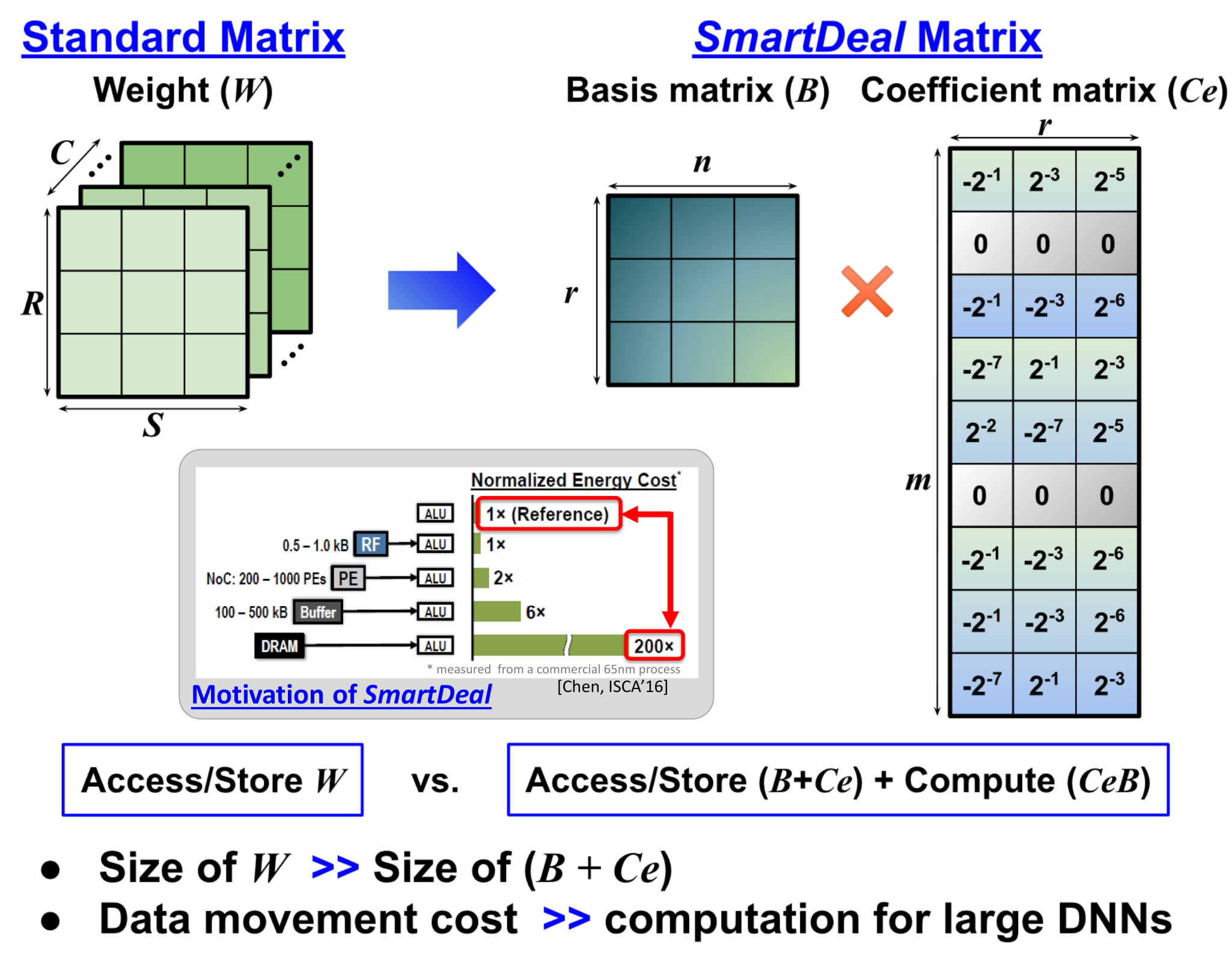}
\vspace{-1em}
\caption{Overview of the proposed weight re-structuring  in \textit{SmartDeal}.}
\vspace{-1em}
\label{SmartDeal}
\end{figure}
However, the excellent performance comes at a heavy parameter cost, which needs external dynamic random-access memory (DRAM) for storage. The prohibitive energy of DRAM accesses makes DNN deployment on resource-constrained devices non-trivial. Take the accelerator in~\cite{chen2014diannao} as an example: $>95$\% of the energy is consumed by DRAM.
Thus, it is crucial to minimize weights and data movements in order to improve the energy efficiency of DNNs.

We present a holistic \textbf{algorithm-hardware co-design} framework, called \textit{SmartDeal}, to aggressively reduce both the energy consumption of data movement and the storage for weights: the two major limiting factors for DNN on-device deployment. Our \textbf{underlying philosophy} is to seek a ``smart deal'': \textit{trading the more ``expensive'' memory storage/access for ``cheaper'' computation, to eliminate the dominant data movement cost}. Our \textbf{technical contributions} highlight three major aspects: (1) \textbf{inference algorithm}: whose core is a novel \textit{SmartDeal} weight representation and a unified optimization framework; (2) \textbf{inference hardware accelerator}, which is co-designed to maximize the benefit of our inference algorithm; and (3) \textbf{training algorithm}, which extends the \textit{SmartDeal} benefits to energy-efficient training in non-trivial ways. We will go through each of them with more details in Section~\ref{gothrough}.

The paper significantly extends our previous conference version~\cite{zhao2smartexchange}. \underline{First and Foremost}, our prior work \cite{zhao2smartexchange} only covers the inference algorithm and its hardware accelerator, while this current work proposes \textit{SmartDeal}-compatible training techniques for the first time, exploring the new horizon of energy-efficient training. We also extensively benchmark the efficient training performance, in two common edge-based training settings (adaptation and fine-tuning), using a training-specific hardware accelerator. \underline{Second}, we offer a comprehensive suite of ablation studies to carefully investigate the impact of each proposed component in the \textit{SmartDeal} weight decomposition. \underline{Besides}, more datasets and application scenarios are included in experiments, such as a language modeling dataset.

\subsection{Overview of Major Technical Contributions}
\label{gothrough}
We start by introducing \textit{SmartDeal} for \textbf{inference}, whose core is a new weight representation: a layer-wise weight matrix is decomposed as the product of a small \textit{basis matrix} and a large \textit{coefficient matrix} (see Fig. \ref{SmartDeal}). We simultaneously enforce two important properties on the latter coefficient matrix:
\begin{itemize} \itemsep -1pt
    \item \textit{(Structurally) sparse}: most elements are zero. Besides element-wise sparsity, we also exploit structured sparsity \cite{wen2016learning} for more hardware-friendliness.
    \item \textit{Specially quantized}: the non-zero elements take only power-of-2 values, which not only have compact bit representations, but also (more importantly) turn the dot-product multiplication with the non-zero elements into much lower-cost shift-and-add operations.
\end{itemize}
It is worth emphasizing that we never claim to have invented any new weight factorization, or quantization, or structured sparsity algorithm in this paper. The \textit{essential novelty} of \textit{SmartDeal} lies in its unique angle and motivation, as well as the unified optimization framework dedicated to achieving our goal. The \textit{core theme} of \textit{SmartDeal} is trading higher-cost memory storage/access for lower-cost computation: this is beyond simply compressing weights to save storage or FlOPs. Our weight re-structure could be interpreted as an innovative, well-motivated integration of \textit{sparsification (or pruning)}, \textit{factorization}, and \textit{quantization} on DNN weights, that can be solved under one unified optimization algorithm.

To fully unleash \textit{SmartDeal} algorithm's potential, we further develop a dedicated DNN \textbf{inference accelerator} that takes advantage of the much reduced weight storage and readily-quantized weights resulting from the algorithm to enhance hardware acceleration performance. Experiments show that  the proposed accelerator outperforms state-of-the-art DNN inference accelerators in terms of acceleration energy efficiency and latency by up to 6.7$\times$ and 19.2$\times$, respectively.
 
We then take another big leap to extend \textit{SmartDeal} from inference to DNN \textbf{training}. 
The current practice of edge-based training typically starts from a pre-trained and pre-loaded model, and then continues tuning the model with more data collected from the same training domain (denoted as \textit{fine-tuning}), or from a different new domain (denoted as \textit{adaptation}) due to customization or personalization \cite{e^2_train,you2019drawing,li2020halo}. Although fine-tuning or adaptation costs much less compared to training from scratch,
their large resource consumption stands at odds with the limited computing and energy resources at the edge \cite{Yang_2017}. To enable \textit{SmartDeal} for energy-efficient on-device training, we aim to preserve the \textit{SmartDeal} re-structured weight form during training. To do so, we have to cope with two roadblocks in the resulting optimization:
\begin{itemize} \itemsep -1pt
    \item The weights, and therefore their sparse coefficient matrices, keep changing during training: that was known to be very hardware-unfriendly. To deal with this challenge, we perform a \textit{SmartDeal} decomposition on the pre-trained weight initializations, and then maintain the (structured) sparsity map (i.e., the locations of nonzero elements) unchanged during training, while their magnitudes can be updated. This greatly saves memory access and energy overhead with little impact on achievable performance\footnote{While it is not directly applicable to training from random scratch, such setting is unlikely in resource-constrained training.}.
    \item 
    The other challenge arises from our enforced structure that non-zero coefficients could only take discrete values (power-of-2). The gradient descent cannot be directly applied to the discrete domain; meanwhile it is highly inefficient to keep a copy of continuous/higher-precision latent weight for updating. Instead, we design a special and light-weight update rule for the coefficient matrix, that turns floating-point additions into quantization bucket switches, to be decided by gradient signs \cite{bernstein2018signsgd}.
\end{itemize}
We evaluate \textit{SmartDeal} on one state-of-the-art training accelerator \cite{kim20192} with necessary modifications to demonstrate the generality where \textit{SmartDeal} achieves up to 4.48$\times$ improvement in energy efficiency over state-of-the-art competitors for training, at the negligible accuracy losses.

\section{Background and related work}

\subsection{A Motivating Example for \textit{SmartDeal}}

\begin{scriptsize}
\begin{table}[!t]
  \centering
  \caption{Unit energy cost per 8-bit extracted from a commercial 28nm technology.}
  \label{table:formatting}
  \begin{tabular}{|c|c|c|c|c|c|}
    \hline
     & DRAM & SRAM & MAC & multiplier & adder \\
    \hline
    Energy  & \multirow{2}{*}{100} & \multirow{2}{*}{1.36$-$2.45} &  \multirow{2}{*}{0.143} &  \multirow{2}{*}{0.124} & \multirow{2}{*}{0.019} \\
    (pJ/8bit) & & & & & \\
    \hline
  \end{tabular}
  \label{table:unit_energy}
\end{table}
\end{scriptsize}

Table~\ref{table:unit_energy} shows the unit energy cost of accessing different-level memories with different storage capacities and computing an MAC/multiplication/addition (the main computation operation in DNNs) designed in a commercial 28nm CMOS technology. We can see that the unit energy cost of memory accesses is much higher ($\geq 9.5\times$) than that of the corresponding MAC computation. Therefore, it is promising in terms of more efficient acceleration if we can potentially enforce higher-order of weight structures to more aggressively trade higher-cost memory accesses for lower-cost computations, motivating our \textit{SmartDeal} idea. That is, the resulting higher structures in DNN weights' decomposed matrices, e.g., $C_e$ in Figure~\ref{SmartDeal}, will enable much reduced memory accesses at a cost of more computation operations (i.e., shift-and-add operations in our design), as compared to the vanilla networks. 

\subsection{Basics of Deep Neural Networks}
Modern DNNs usually consist of a cascade of multiple convolutional (CONV), pooling, and fully-connected (FC) layers through which the inputs are progressively processed. The CONV and FC layers can be described as:
\begin{equation}
    \begin{split}
    &\textit{\textbf{O}}[c_o][e][f]=\\
    & \sigma(\sum_{c_i,k_r,k_s}^{C,R,S}{
        \textit{\textbf{W}}[c_o][c_i][k_r][k_s]}\cdot \textit{\textbf{I}}[c_i][eU+k_r][fU+k_s]+\textit{\textbf{B}}[c_i]
    ),\\
    & 0\le c_o < M,~0\le e < E, 0\le f < F,
    \end{split}
    \label{eq:CONV}
\end{equation}
where \textit{$\textbf{W}$}, \textit{$\textbf{I}$}, \textit{$\textbf{O}$}, and \textit{$\textbf{B}$} denote the weights, input activations, output activations, and biases, respectively. In the CONV layers, $C$ and $M$, $E$ and $F$, $R$ and $S$, and $U$ stand for the number of input and output channels, the size of input and output feature maps, and the size of weight filters, and stride, respectively; while in the FC layers, $C$ and $M$ represent the number of input and output neurons, respectively; with $\sigma$ denoting the activation function, e.g., a $ReLU$ function ($ReLU(x)=max(x,0)$). The pooling layers reduce the dimension of feature maps via average or max pooling. The recently emerging compact DNNs (e.g., MobileNet~\cite{howard2017mobilenets} and EfficientNet~\cite{tan2019efficientnet}) introduce depth-wise CONV layers and squeeze-and-excite layers which can be expressed in the above description as well~\cite{chen2019eyeriss}. 

\subsection{Overview of Efficient Deep Learning}
To reduce the large quantity of weight parameters, numerous DNN compression techniques have been proposed to shrink the weight redundancy and accelerate the inference, including matrix decomposition \cite{novikov2015tensorizing, huang2018highly}, quantization \cite{zhou2016dorefa, lee2018unpu,you2020shiftaddnet}, pruning \cite{han2015learning, han2016eie,wen2016learning}, knowledge distillation \cite{ba2014deep,hinton2015distilling,you2020shiftaddnet}, and dynamic inference \cite{wang2018skipnet,shen2020fractional,hu2020triple,wang2020dual}. Combinations of two techniques have also been studied, e.g. decomposition with pruning \cite{Yu_2017_CVPR}, distillation with quantization \cite{mishra2017apprentice,polino2018model}, and distillation with pruning \cite{wang2018graph}. Some latest works \cite{gui2019adversarially,wang2020gan} start to combine and jointly optimize three compression ideas, but for different motivations and applications (e.g., compressing robust models, and GANs). To our best knowledge, 
\textit{SmartDeal} features a new joint formulation that combines the three ideas of weight pruning, matrix decomposition, and power-of-2 quantization: a unification never being considered by peer works.
Also differently from \cite{gui2019adversarially,wang2020gan}, \textit{SmartDeal} is the first to associate with hardware co-design, especially with the unique goal to reduce the memory/storage-access costs. It is also the first of its kind to extend to efficient training.

Besides combining and jointly optimizing multiple compression techniques, another direction of work focuses on improving the performance of quantized models by utilizing more flexible quantization schemes. Mixed-precision training methods \cite{zhou2016dorefa,micikevicius2018mixed,wu2018mixed} quantize weights, activations and gradients to different precisions instead of using single precision for the whole model.
Quantization-aware training (QAT) co-optimizes the quantization schemes and the model parameters \cite{Wang_2019_CVPR,tailor2021degreequant}.
\textit{SmartDeal} is orthogonal to mixed-precision and QAT methods, which can be easily integrated into \textit{SmartDeal} to find its optimal precision configuration.

Moreover, current DNNs are typically trained in resource-rich servers or data centers.
Nevertheless, we see a growing necessity for the model to continue learning and updating itself in situ, such as for user personalization, or incremental/lifelong learning in open-ended environments. On-device local learning can avoid communication forth-and-back between data centers and devices, reducing system latency and enhancing privacy protection. Despite a number of recent efforts \cite{e^2_train,you2019drawing,li2020halo,fu2020fractrain}, limited progress has been witnessed so far in this field, partially due to the even larger gap between the training resource demands and the available on-device resources.

\subsection{Compression-Aware DNN Accelerators}

In general, the three typical compression approaches, weight \textit{factorization}, data \textit{quantization}, and weight \textit{sparsification}, have been exploited in DNN accelerators design to boost energy efficiency.
\cite{huang2018highly} demonstrate DNNs with tensorized factorization using ASIC. 
For the weight sparsification accelerators, \cite{parashar2017scnn,zhang2016cambricon,han2016eie} have been proposed.
Quantization is widely used by inference accelerators \cite{parashar2017scnn, zhang2016cambricon,huang2018highly,han2016eie}. 
In comparison, our proposed \textit{SmartDeal} inference accelerator also unifies three techniques to simultaneously shrink the memory footprint and simplify the computations when recovering the weight matrix during runtime.

\section{\textit{SmartDeal} for Efficient Inference}
\label{sec:inference}

In this section, we introduce the weight decomposition structure in \textit{SmartDeal}, as can be naturally applied to \textit{feed-forward inference} to reduce energy and time consumption of data movement as well as storage.

\subsection{Problem Formulation}
\label{subsec:formulation}

Given a weight matrix $W\in\mathbb{R}^{m\times n}$, we seek to decompose it as the product of a coefficient matrix $C_e\in\mathbb{R}^{m\times r}$ and a basis matrix $B\in\mathbb{R}^{r\times n}$ where $r \le \min\{m,n\}$, such that
\begin{align}
    W \approx C_e B.
    \label{eq:rank}
\end{align}

In practice, $n$ is usually set to be very small (thus small $B$), and $m\gg n$ (thus much larger $C_e$).
Here we assume a 2D weight matrix in a FC layer as example for the simplicity of notation. We will later show that \textit{SmartDeal} algorithm can be easily applied to weight tensors in CONV layers.

In addition to suppressing the reconstruction error (often defined as $||W-C_e B||_F^2$), we expect the decomposed matrix factors to display more favorable structures for compression/acceleration. 
For the much larger $C_e$, we enforce the following two structures simultaneously:
\emph{(i)} $C_e$ needs to be highly sparse (a typical goal of pruning); and
\emph{(ii)} the non-zero elements in $C_e$ are exactly the powers of $2$, so that their bit representations can be very compact and their involved multiplications to rebuild the original weights from $B$ and $C_e$ are simplified into extremely cheap shift-and-add operations.
As a result, instead of storing the whole weight matrix, the new structure requires storing only a very small $B$; and a large, yet highly sparse and readily quantized $C_e$.
We call this process \textit{SmartDeal} (\textbf{SD}) decomposition and the resulting \{$C_e$, $B$\} pair the \textit{SD form} of $W$.

\textit{SmartDeal} decomposition hence can be written as the following constrained optimization:
\begin{align}
\mathrm{arg}\:{\mathrm{min}_{C_e,B}} & \quad\|W - C_e B\|_F^2 \nonumber \\
\mathrm{subject\ to} & \quad {\sum}_j\ \|C_e[:,j]\|_0 \le S_c, \label{eq:sed}\\
& {C_e}[i,j]\in\Omega_P,\ \ \forall i,j, \ |P| \leq N_p \nonumber
\end{align}
where $\Omega_P\coloneqq\{0, \pm 2^p | p\in P\}$ with $P$ being a chosen integer set that includes the possible degrees of the power-of-2 numbers and has cardinality no more than $N_p$, i.e. $|P| \le N_p$.
In Eq.~(\ref{eq:sed}), $S_c$ controls the total number of non-zero elements in $C_e$, i.e. the sparsity, while $N_p$ controls the bit-width required to represent a  element in $C_e$. An innovative assumption of SD is to require non-zero elements in $C_e$ to take one of a few \textit{pre-defined, specifically-picked discrete values}.
That is different from previous compression using weight clustering, whose quantized values are adaptively learned from data~\cite{gong2014compressing,wu2018deep}. This special design is to facilitate (1) compact storage, and more crucially (2) extremely cheap weight reconstruction using only shift-and-add operations - the latter cannot be fulfilled by other arbitrary quantization.

\subsection{\textit{SmartDeal} Decomposition Algorithm}
\label{subsec:alg}

Solving Eq.~(\ref{eq:sed}) is non-trivial due to the nonconvex and integer set constraints. We propose a coordinate descent-type algorithm that iterates between objective fitting and feasible set projection, as outlined in Algorithm~\ref{SD}, and the explanation of three key steps to be iterated are discussed follows.
Here, $\delta(C_e)$ is the difference between two iterates of $C_e$, $tol$ is a small number indicating the convergence of the algorithm, and $max\_iter$ is the max number of iterations.

\begin{algorithm}[H]
    \caption{\textit{SmartDeal} decomposition algorithm.}
    \label{SD}
    \begin{algorithmic}[1]
        \STATE{Initialize $C_e$ and $B$; k = 0}
        \STATE{While \textit{$\|\delta(C_e)\| \geq tol$} or \textit{ k $<$ max\_iter}:}
            \STATE{\hspace{10pt} \textbf{Step 1:} Quantizing $C_e$ to powers of $2$;}
            \STATE{\hspace{10pt} \textbf{Step 2:} Fitting $B$ and $C_e$;}
            \STATE{\hspace{10pt} \textbf{Step 3:} Promoting (structured) sparsify $C_e$;}
            \STATE{\hspace{10pt} k = k + 1;}
            \STATE{Re-quantize $C_e$ and re-fit $B$.}
    \end{algorithmic}
\end{algorithm}

We empirically find that simply initializing $C_e=W$ and $B=I$ can produce robust and good performing decomposition results. Hence, we use this initialization in all experiments. After the initialization, we iteratively perform the following three steps to gradually obtain better decompositions.

\textbf{Step 1: Quantizing $C_e$.} The quantization step projects the nonzero elements in $C_e$ to $\Omega_P$. Specifically, we will first normalize each column in $C_e$ to have a unit norm in order to avoid scale ambiguity. We will then round each non-zero element to its nearest power-of-two value. We define $\delta(C_e)$ to be the quantization difference of $C_e$.

\textbf{Step 2: Fitting $B$ and $C_e$.} We will first fit $B$ by solving
$\mathrm{arg}\:\mathrm{min}_B  \|W - C_e B\|_F^2$,
and then fit $C_e$  by solving
$\mathrm{arg}\: \mathrm{min}_{\,C_e}  \|W - C_e B\|_F^2$.
When fitting either one, the other is fixed to be its current updated value. The step simply deals with two unconstrained least squares.

\textbf{Step 3: Sparsifying $C_e$.} We then prune the non-zero elements in $C_e$ with smallest magnitudes to promote more sparsity. In practice, we use hard thresholds for element-wise and structured sparsity to zero out small magnitudes in $C_e$ for implementation convenience. When promoting element-wise sparsity in $C_e$, as sparsity level $S_c$ usually needs to be manually adjusted for each layer, we instead use a heuristic threshold $\theta$ to zero out elements. The structured sparsity is discussed in details in \ref{sec:structure_sparsity}.

After sufficiently iterating between the above three steps (i.e., quantization, fitting and sparsification), we conclude the iterations by re-quantizing the nonzero elements in $C_e$ to ensure ${C_e}[i,j]\in\Omega_P$ and then re-fitting $B$ with the updated $C_e$. An example of how the $C_e$ and $B$ matrices evolve along the iterations is given in Appendix~\ref{sec:evolution}.

\subsection{Applying the SmartDeal Algorithm to DNNs}
\label{subsec:sd2dnn}

\subsubsection{\textit{SmartDeal} algorithm as post-processing}
The selection of the dimensions of the coefficient matrix $C_e$ and the basis matrix $B$ is a design knob of \textit{SmartDeal} for trading-off the achieved \textit{compression rate} and model accuracy, i.e., a smaller $r$ (see notations in \ref{subsec:formulation}) favors a higher compression rate yet might cause a higher accuracy loss. Note that
$r$ is equal to the rank of the basis matrix $B$, i.e., $r=n$ when $B$ is a full matrix, otherwise $r\leq n$.
To minimize the memory storage, we set the basis matrix $B\in\mathbb{R}^{r\times n}$ to be small. 
In practice, we choose $n=R=S$ with $R\times S$ being the CONV kernel size. Since $n$ is small, we choose $r = n = S$ too. 
We next discuss applying the proposed algorithm to the FC and CONV layers. 
\begin{itemize}
    \item \underline{\textit{\textit{SmartDeal} on FC layers.}} Consider a fully-connected layer $W \in R^{M \times C}$. We reshape each row of $W$ into a new matrix $\tilde{W_i}\in R^{C/S \times S}$, and then apply \textit{SmartDeal} algorithm. Specifically, zeros are padded if $C$ is not divisible by $S$, and 
    \textit{SmartDeal} algorithm is applied to $\tilde{W_i}$, where $i=1,\dots,M$. When $C\gg S$, the reconstruction error might tend to be large due to the imbalanced dimensions. We alleviate it by slicing $\tilde{W_i}$ into smaller matrices along the first dimension.
    \item \underline{\textit{\textit{SmartDeal} on CONV layers.}} Consider a convolutional layer $W$ in the shape $(M, C, R, S)$.
    \underline{Case 1}: $R=S>1$. We reshape the $M$ filters in $W$ into matrices of shape $(S \times C, S)$, on which \textit{SmartDeal} algorithm is applied. The matrices can be sliced into smaller matrices along the first dimension if $S \times C\gg S$.
    \underline{Case 2}: $R=S=1$. The weight is reshaped into a shape of $(M,C)$ and then is treated the same as an FC layer.
\end{itemize}
The above procedures are easily parallelized along the axis of the output channels for acceleration.

Applying \textit{SmartDeal} algorithm to a VGG19 network\footnote{https://github.com/chengyangfu/pytorch-vgg-cifar10} pre-trained on the CIFAR-10 \cite{cifar10}, with $\theta=4\PLH10^{-3}$, $tol=10^{-10}$ (introduced in Section~\ref{subsec:alg}), and a maximum iteration of 30, the accuracy drop in the validation set is as small as $3.21\%$ with an overall compression rate of over 10$\times$ without re-training after the decomposition. The \textit{overall compression rate} of a network is defined as the ratio between the total number of bits to store the weights (including the coefficient matrix $C_e$, basis matrix $B$, and encoding overhead) and the number of bits to store the original FP32 weights.

\subsubsection{Enhancing Accuracy with Re-Training}

After a DNN has been post-processed by \textit{SmartDeal} algorithm, a re-training step can be used to remedy the accuracy drop. As the un-regularized re-training will break the desired property of coefficient matrix $C_e$, we take an empirical approach to alternate between 1) re-training the DNN for one epoch; and 2) applying \textit{SmartDeal} algorithm to ensure the $C_e$ structure. The default iteration number is 50 for CIFAR-10 \cite{cifar10} and 25 for ImageNet \cite{Deng09imagenet}. As shown in the ablation experiments in \ref{subsec:compression_results}, the alternating re-training process improves the accuracy while maintaining the favorable weight structure. More analytic solutions will be explored in future work, e.g., incorporating \textit{SmartDeal} algorithm as a regularization term \cite{wu2018deep}. 

\subsubsection{Time complexity of SmartDeal} \textit{SmartDeal} algorithm is efficient in terms of running time with fully parallelized implementation. In practice, running one pass of the parallelized SmartDeal algorithm takes less than 30 seconds for VGG-19 and ResNet-18 networks and less than 2 minutes for ResNet-50. In combined with the alternating re-training approach mentioned above, the computational cost of \textit{SmartDeal} is negligible compared to the cost for training on data. The above running time is measured on Intel Xeon Platinum 8168 CPU platform (we only implement \textit{SmartDeal} on CPU currently).

\subsection{\textit{SmartDeal} with Structured Sparsity}
\label{sec:structure_sparsity}

Customized accelerators for sparse models can utilize the sparsity to reduce the associated computations and memory accesses \cite{parashar2017scnn,zhang2016cambricon}. However, the irregularity in element-wise/unstructured sparse models prevents hardware from fully leveraging reduction. Coarse-grained sparsity brings more regular sparsity pattern, making it easier for hardware acceleration \cite{mao2017exploring}.
Hereby, when we design the dedicated hardware accelerator for \textit{SmartDeal}, we introduce two types of structured sparsity, \textit{channel-wise} and \textit{vector-wise} sparsity, to $C_e$:
\begin{itemize}
    \item We first prune channels whose corresponding scaling factor in batch normalization layers is lower than a threshold which is manually controlled for each layer. In practice, we only apply channel-wise sparsifying at the first training epoch once, given the observation that the pruned channel structure will not change much. 
    \item We then zero out elements in $C_e$ based on the magnitudes to meet the vector-wise sparsity constraint: $\sum_j\ \|C_e[:,j]\|_0 \le S_c$, where $S_c$ is manually controlled per layer. 
\end{itemize}
Structured sparsity being more regular and hardware efficient, it also brings much more aggressive constraints on the model capacity and thus more performance degradation. This is empirically verified in the ablation study in Section~\ref{sec:exp-inference-ablation}.
Although vector-wise sparsity has been investigated in \cite{yu2017scalpel}, we are the first to consider vector sparsity in a uniﬁed framework with quantization, sparsification and decomposition.

After we obtain $C_e$ and $B$ via \textit{SmartDeal} algorithm, we further desire storage-economic and hardware-friendly representations to store them on-device. We explain the encoding schemes of $C_e$ and $B$ in Appendix \ref{subsec:encoding}. We also discuss more on the rationale behind our algorithm, in Appendix \ref{subsec:sd-discussions}.

\section{\textit{SmartDeal} for Efficient Training}
\label{sec:SmartDeal_Training}

Extending \textit{SmartDeal} to energy-efficient training is highly non-straightforward. As pointed out in Section \ref{sec:introduction}, the dynamic sparsity pattern and the discrete values in $C_e$ constitute two grand challenges. Also, directly involving optimization like solving Eq.~(\ref{eq:sed}) into training is not acceptable due to its own complexity.
We hereby discuss how to transplant the methodology of \textit{SmartDeal} to energy-efficient training, with two dedicated techniques presented to address the two challenges.
The two key points of the proposed techniques are (1) to optimize the decomposed $C_e$ and $B$ matrices in the special discrete space constrained by sparsity and power-of-2 quantization so that the model preserves \textit{SmartDeal} structure and thus the economic memory access throughout training; (2) to avoid using any ``shadow weights'' i.e., the latent high-precision copies of weights that are actually trained and will introduce large overhead in model storage.

\textbf{Basic routine: SD-Training (SD-T).} Considering the practice of on-device learning, we assume a pre-trained DNN to start with, where the goal is either fine-tuning or adaptation.
For a layer with pre-trained weight $W$, we first perform one-pass SD to get the initial $C_e$ and $B$. We use $C_e$ and $B$ as hidden weights that will be used to reconstruct $W$ run-time during a \textit{feed-forward} pass.
Note that the reconstruction step introduces little overhead but significantly lowers the data movement cost due to the fixed sparsity structure and the power-of-2 non-zero values in $C_e$.

During the \textit{back-propagation pass}, the gradient is passed back to the intermediate $W$ as usual. The gradients of $B$ and $C_e$ are calculated with $W$'s gradient and matrix multiplications. We update $B$ using the standard gradient descent. For updating $C_e$, we explicitly require $C_e$ to be within its original feasible domain: a (structurally) sparse matrix with power-of-2 non-zero elements. Such a challenging requirement is met thanks to the next two customized techniques. 

\textbf{\#1. For $C_e$ zeros: fixed sparsity mask.} We fix the sparsity pattern in $C_e$ throughout training: the initial zero entries in $C_e$ are ``frozen'' to zero, while the initial non-zero entries can be updated to either zero or non-zero flexibly. Experiments show that such a fixation does not noticeably impact the tuned/adapted model accuracy. 
This fixed sparsity brings in two-fold advantages:
1) fixing the sparse pattern of $C_e$ saves the gradient computations of its zero elements during training, which can not be skipped in classic pruning with a dynamic sparse pattern;
2) the static sparsity pattern can be utilized to avoid the dynamic indexing sparse weights and/or adjusting processing schedules. Extensive experiments in Tab. \ref{table:sd-t-c100} show that the fixed sparsity implementation is effective across different models and datasets for saving more energy.

\textbf{\#2. For $C_e$ nonzeros: bucket switch updating.} The next dilemma is on updating non-zero power-of-2 elements: if we update $C_e$ using floating-point add operations, the floating-point numbers have to be recovered before updating incurring overheads. Instead, we propose a \textit{Bucket Switch} updating scheme for $C_e$ non-zero updating: inspired by \cite{bernstein2018signsgd} showing that taking only gradient signs (i.e., 1-bit gradients) suffices to training DNNs, we refer to gradient signs to guide the switch of non-zero values, from one discrete ``bucket'' to another. 

Specifically, if the gradient w.r.t. a non-zero element (whose current value is $2^p$) is positive, then we will switch up its value from $2^p$ to $2^{p+1}$ (or no change if $p = P$ already reaches the upper range bound). Similarly, a negative gradient will switch $2^p$ down to $2^{p-1}$, and a zero gradient will not change it. In this way, the non-zero element in $C_e$ is directly updated over the discrete domain, without any overhead of floating-point number operations. Notice that, we never aggregate high-precision updates, and there also exists no high-precision latent weight for $C_e$ in our implementation, because such will go against our goal of saving storage and energy for efficient training. Instead, we only record update directions (signs) and accumulate using an \textit{integer} counter. Once the counter's integer record passes a threshold, we ``switch the bucket". 

While the above gradient quantization rule could suffer from high variance to learning rates, in practice, we apply two methods to mitigate the effect of noisy gradients.
Before taking the sign, gradients whose magnitudes are below a threshold $\theta_g$ are zeroed out.
We also adopt an ``update delay and aggregation'' scheme, in which we only really update an entry in $C_e$ (switch the bucket) when it receives switch signals in the same direction for enough times. The number of required times is controlled by an integer hyperparameter $\theta_c$.
That is inspired by the lazy update and trajectory smoothing in gradient-based optimization \cite{izmailov2018averaging,yang2019swalp}: applying the similar idea to bucket switch is found to help training stability (as our update directions are ``noisy'').
The detailed description of this strategy is laid out in Appendix~\ref{sec:bucket-switch}.
We note that a similar idea of ``bucket switch'' and ``update decay and aggregation'' was coincidentally found useful in another latest work on optimizing binary neural networks (BNNs) \cite{helwegen2019latent}. We leave more discussions on the potential theoretical underpinnings for future work.
We also apply the stochastic weight averaging (SWA) technique \cite{izmailov2018averaging} that could stabilize training too without incurring noticeable overhead.

\section{Dedicated Hardware-Algorithm Codesign}
\label{sec:Dedicated_accelerator}

In this section, we present our proposed \textit{SmartDeal} accelerator. We first introduce the design principles and considerations (Section \ref{sec:Design_challenges_principles}) for fully making use of the proposed \textit{SmartDeal} algorithm's properties to maximize energy efficiency and minimize latency, and then describe the proposed accelerator (Section \ref{subsec:accelerator}) in details.

\subsection{Design Principles and Considerations}
\label{sec:Design_challenges_principles}

\textbf{Minimizing overhead of rebuilding weights. }
Thanks to the sparse and readily quantized coefficient matrices resulting from the \textit{SmartDeal} algorithm, the memory storage and data movements associated with these matrices can be greatly reduced (see Table \ref{table:sed_results}; e.g., up to $80\times$). Meanwhile, 
to fully utilize the advantages of the \textit{SmartDeal} algorithm, the overhead of rebuilding weights should be minimized. To do so, it is critical to ensure that the location and time of the rebuilding units and process are properly designed. Specifically, for a \textit{SmartDeal} accelerator 
1) the rebuild engine (RE) that restores weights using both the basis matrix and corresponding weighted coefficients should be located closes to the PEs for minimizing the data movement cots of the rebuilt weights. 
2) As the basis matrices are reused most frequently, the dataflow for these matrices should be weight stationary, i.e., once being fetched from the memories, they stay in the REs until all the corresponding weights are rebuilt.

\begin{figure}[!t]
   \vspace{-1em}
    \centerline{\includegraphics[width=83mm]{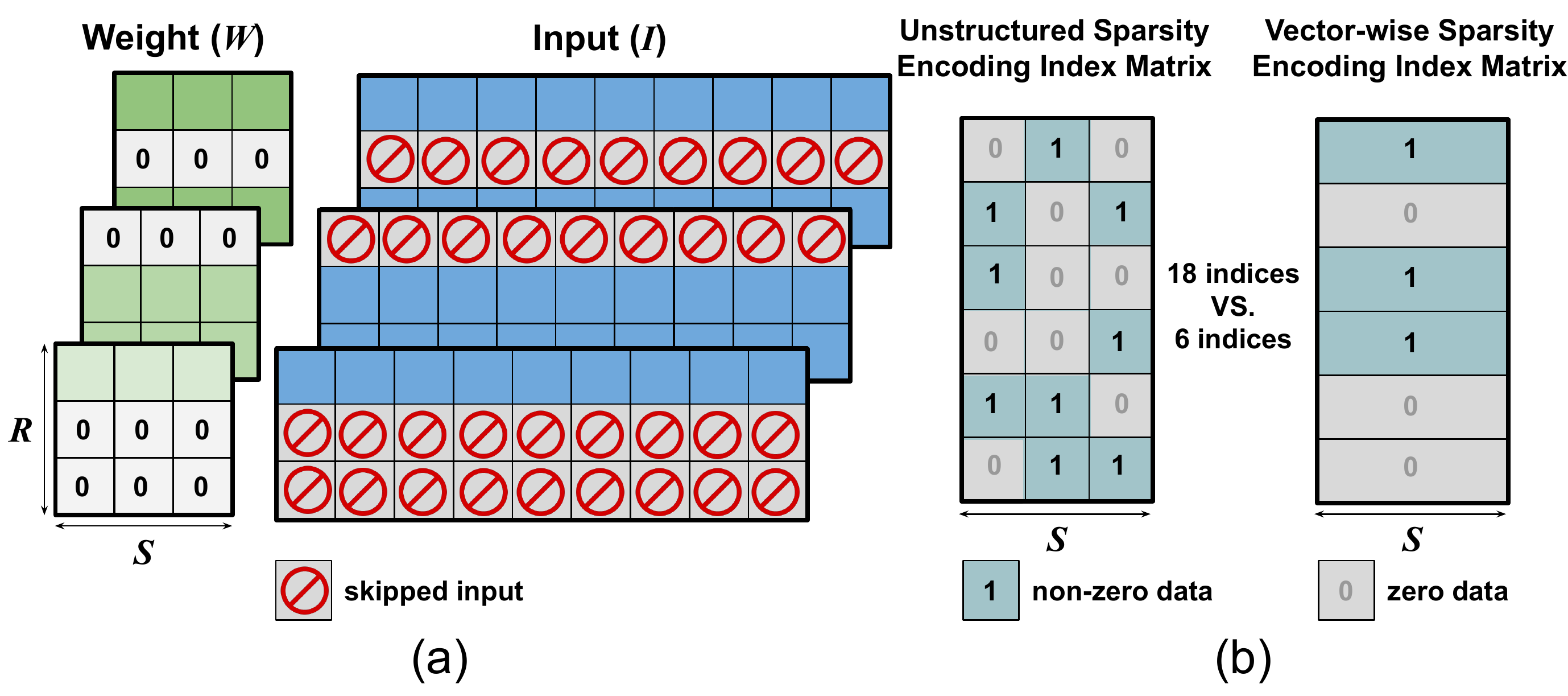}}
    \vspace{-1em}
    \caption{An illustration of (a) vector-wise skipping the corresponding activations, and (b) the reduced indexing overhead, thanks to the enforced \textit{vector-wise} weight sparsity of the \textit{SmartDeal} algorithm.
       \vspace{-1em}
    \label{fig:coefficient_sparsity}}
\end{figure}

\textbf{Taking advantage of the structured sparsity. } The enforced \textit{vector-wise} sparsity in the \textit{SmartDeal} algorithm's coefficient matrices offers benefits of 1) \textit{vector-wise} skipping both the memory accesses and computations of the corresponding activations (see Figure \ref{fig:coefficient_sparsity} (a)) and 2) reduced coefficient matrix encoding overhead (see Figure \ref{fig:coefficient_sparsity} (b)). Meanwhile, there is an opportunity to make use of the \textit{vector-wise/bit-level} sparsity of activations for improving efficiency.

\begin{figure}[h]
   \vspace{-1em}
    \centerline{\includegraphics[width=88mm]{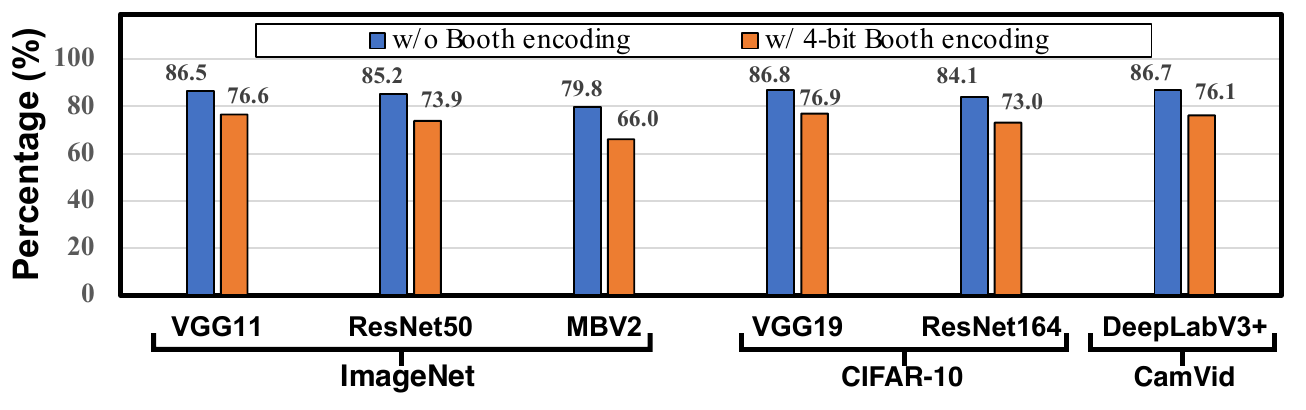}}
       \vspace{-1em}
    \caption{The bit-level sparsity in activations for six models on three datasets.} 
       \vspace{-1em}
    \label{fig:activatioin_sparsity}
\end{figure}

\ul{First}, one promising benefit of the \textit{SmartDeal} algorithm's enforced \textit{vector-wise} sparsity in the coefficient matrices is the possibility to \textit{vector-wise} skip both the memory accesses and computations of the corresponding activations (see Figure \ref{fig:coefficient_sparsity} (a)). This is because those \textit{vector-wise} sparse coefficient matrices' corresponding weight vectors naturally carry their \textit{vector-wise} sparsity pattern/location, offering the opportunity to directly use the sparse coefficient matrices' encoding index to identify the weight sparsity and skip the corresponding activations' memory accesses and computations. Such a skipping can lead to large energy and latency savings because weight vectors are shared by all activations of the same feature maps in CONV operations, see Figure \ref{fig:coefficient_sparsity} (b).

\begin{figure*}[!t]
   \vspace{-1em}
    \centerline{\includegraphics[width=168mm]{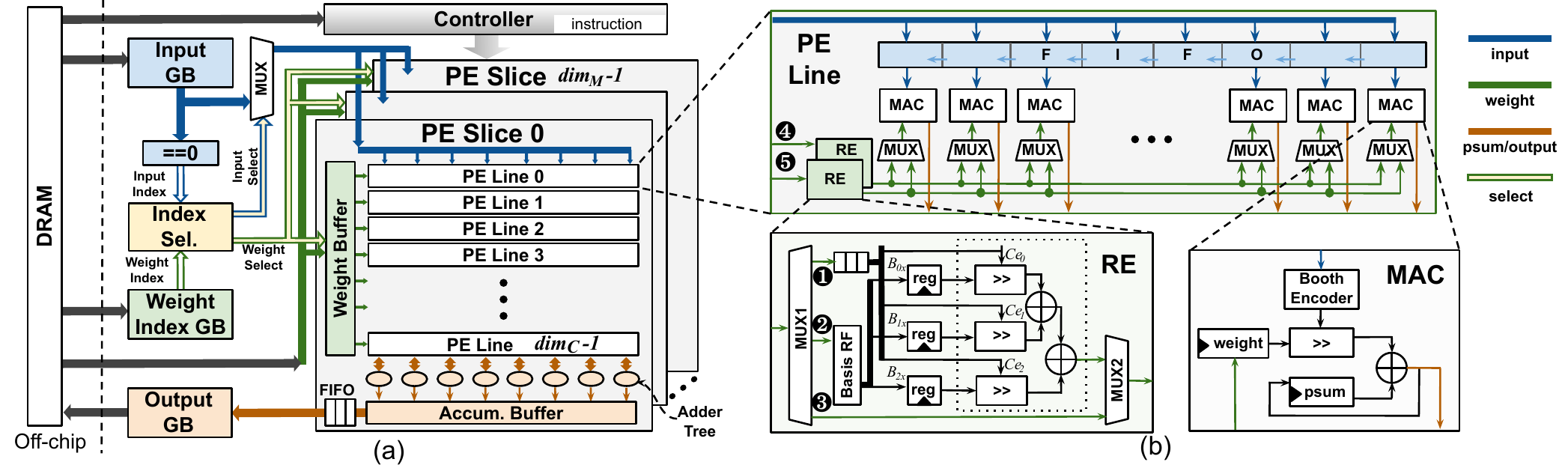}}
       \vspace{-1em}
    \caption{An illustration of the proposed \textit{SmartDeal} accelerator: (a) architecture, and (b) the block diagram of the processing element (PE) line, each of which includes two rebuilding engines (REs) and eight multiply-and-accumulate (MAC) units.}
       \vspace{-1em}
    \label{fig:overview_PE}
\end{figure*}

\ul{Second}, commonly used methods for encoding weight sparsity, such as the 1-bit direct weight indexing ~\cite{zhou2018cambricon}, Compressed Row Storage (CRS)~\cite{han2016eie}, and Huffman encoding, store both the values and sparsity encoding indexes of weights. Our \textit{SmartDeal} algorithm's \textit{vector-wise} weight sparsity reduce both the sparsity encoding overhead (see Figure \ref{fig:coefficient_sparsity} (b)) and skipping control overhead. The resulting energy and latency benefits depend on the sparsity ratio and pattern, and hardware constraints (e.g., memory bandwidths).

\ul{Third}, the accelerator can further make use of \textit{bit-level }and \textit{vector-wise} sparsity of activations to  improve energy efficiency and reduce latency, where the \textit{bit-/vector-wise} sparsity means the percentage of the zero activation bits/rows over the total activation bits/rows. 
Figure \ref{fig:activatioin_sparsity} shows the \textit{bit-level} sparsity of activations w/ and w/o 4-bit Booth encoding~\cite{delmas2019bit} in popular DNNs, including VGG11, ResNet50, and MobileNetV2 on ImageNet, VGG19 and ResNet164 on CIFAR-10, and DeepLabV3+ on CamVid. We can see that the bit-level sparsity is 79.8\% under an 8-bit precision and 66.0\% using the corresponding 4-bit Booth encoding even for a compact model like MobileNetV2; for vector-wise sparsity, it can be widely observed among the CONV layers with $3 \times 3$ kernel size, e.g., up to 27.1\% in the last several CONV layers of MobileNetV2 and up to 32.4\% in ResNet164.

\textbf{Support for compact models. } The recently emerged compact models, such as MobileNet \cite{howard2017mobilenets} and EfficentNet \cite{tan2019efficientnet}, often adopt depth-wise CONV and squeeze-and-excite layers other than the traditional 2D CONV layers to restrict the model size, which reduces the data reuse opportunities. Taking a depth-wise CONV layer as an example, it has an ``extreme" small number of CONV channels (i.e., 1), reducing the input reuse over the standard CONV layers; for squeeze-and-excite, similar to that of FC layers, there are no weight reuse opportunities in squeeze-and-excite layers. On-device efficient accelerators should consider these features of compact models for their wide adoption and leveraging compact models for more efficient processing.

\subsection{Architecture of the SmartDeal Accelerator}
\label{subsec:accelerator}

\textbf{Architecture overview.} Figure~\ref{fig:overview_PE} (a) shows the architecture of the proposed \textit{SmartDeal} accelerator which consists of a 3D PE array with a total of $dim_M$ PE slices, input/index/output global buffers (see the blocks named Input GB, Weight Index GB, and Output GB, where GB denotes global buffer) associated with an index selector for sparsity (see the blocks named Index sel.), and an controller. The accelerator communicates with an off-chip DRAM through DMA (direct memory access) \cite{zhang2016cambricon}. Following the aforementioned design principles and considerations (see Section \ref{sec:Design_challenges_principles}), the proposed accelerator features the following properties: \ul{1) \textit{an RE design}} which is inserted within PE lines to reduce the rebuilding overhead (see the top part of Figure~\ref{fig:overview_PE} (b)); 
\ul{2) \textit{a hybrid dataflow}}: an 1D row stationary dataflow is adopted within each PE line for maximizing weight and input reuses, while each PE slice uses an output stationary dataflow for maximizing output partial sum reuses; 
\ul{3) \textit{an index selector}} (named Index Sel. in Figure~\ref{fig:overview_PE} (a)) to select the none-zero coefficient and activation vector pairs as inspired by~\cite{zhou2018cambricon}. 
This is to skip not only computations but also data movements associated with the sparse rows of the coefficients and activations. The index selector design in \textit{SmartDeal} is the same as that of \cite{zhou2018cambricon} except that Huffman encoding is used here and the index values of 0/1 stand for vector (instead of scalar) sparsity; 
\ul{4) \textit{a data-type driven memory partition}} in order to use matched bandwidths (e.g., a bigger bandwidth for the weights/inputs and a smaller bandwidth for the outputs) for different types of data to reduce the unit energy cost of accessing the SRAMs which is used to implement the GB blocks~\cite{du2015shidiannao}. We adopt separated centralized GBs to store the inputs, outputs, weights and indexes, respectively, and distributed SRAMs (see the Weight Buffer unit in Figure~\ref{fig:overview_PE} (a)) among PE slices to store weights (including the coefficients and basis matrices); 
and \ul{5) \textit{a bit-serial multiplier based MAC array}} in each PE line to make use of the activations' bit-level sparsity together with a Booth Encoder as inspired by~\cite{delmas2019bit}.

\begin{figure}[!t]
    \centerline{\includegraphics[width=60mm]{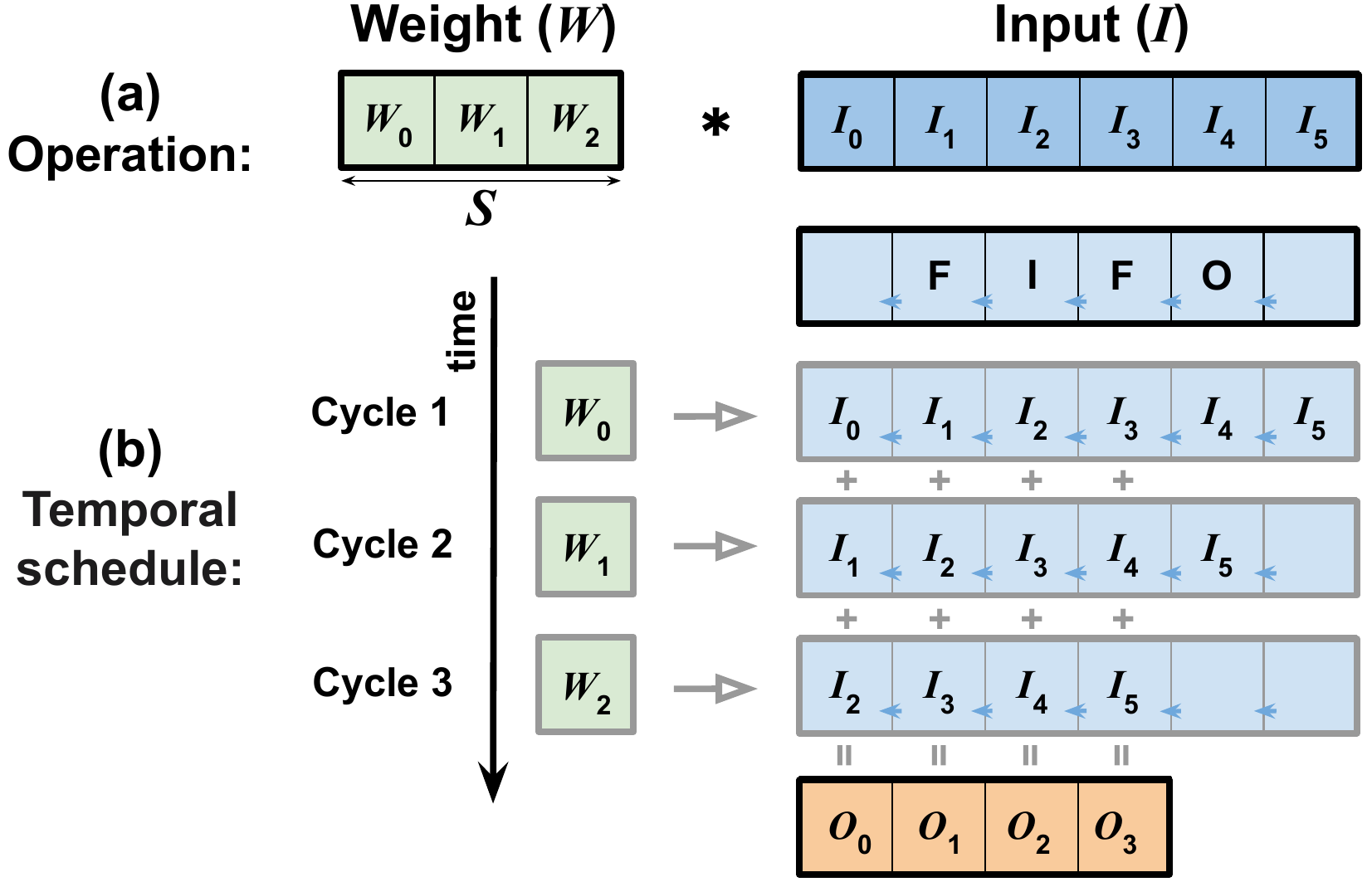}}
    \vspace{-0.5em}
    \caption{An illustration of the proposed 1D row stationary along each PE slice (in this particular example, FIFO size is 5, and in general it should be $dim_F+S-1$): (a) 1D CONV and (b) processing flow of 1D row stationary. }
    \vspace{-1em}
    \label{fig:1D_conv}
\end{figure}

\textbf{PE slices and dataflow.} We here describe the design of the PE Slice unit in the 3D PE slice array of Figure~\ref{fig:overview_PE} (a): 

\ul{First, the 3D PE Slide array:} our \textit{SmartDeal} accelerator enables paralleled processing of computations associated with the same weight filter using the PE slice array of size $dim_M$ (with each PE slice having $dim_C$ PE lines) and $dim_C$ number of input channels, where the resulting partial sums are accumulated using the adder trees at the bottom of the PE lines (see the bottom right side of Figure~\ref{fig:overview_PE} (a)). In this way, a total of $dim_M$ consecutive output channels (i.e., $dim_M$ weight filters) are processed in parallel to maximize the reuse of input activations. Note that this dataflow is employed to match the way we reshape the weights as described in Section~\ref{sec:inference}.C.

\ul{Second, the PE line design:} each PE line in Figure~\ref{fig:overview_PE} includes an array of $dim_F$ MACs, one FIFO (using double buffers), and two RE units, where the REs at the left restore the original weights in a row-wise manner. 
During operations, each PE line processes one or multiple 1D CONV operations, similar to the 1D row stationary in \cite{eyeriss} except that we stream each rebuild weight of one row temporally along the MACs for processing one row of input activations. 
In particular, the 1D CONV operation is performed by shifting the input activations along the array of MACs within the PE line (see Figure~\ref{fig:1D_conv}) via an FIFO; this 1D CONV computation is repeated for the remaining 1D CONV operations to complete one 2D CONV computation in $\leq (S\times R$) cycles (under the assumption of w/ sparsity and w/o bit-serial multiplication) with 1) each weight element being shared among all the MACs in each cycle, and 2) the intermediate partial sums of the 2D CONV operations are accumulated locally in each MAC unit (see the bottom right part of Figure~\ref{fig:overview_PE} (b)). 

\ul{Third, the RE design:} as shown in the bottom left corner of Figure~\ref{fig:overview_PE} (b), an RE unit includes an RF (register file) of size $S\times S$ to store one basis matrix and a shift-and-add unit to rebuild weights. The time division multiplexing unit at the left, i.e., MUX1, is to fetch the {\color{black}{\ding{182}}} coefficient matrices, {\color{black}{\ding{183}}} basis matrices, or {\color{black}{\ding{184}}} original weights. This design enables the accesses of these three types of data to be performed in a time division manner in order to reduce the weight bandwidth requirement by taking advantage of the fact that it is not necessary to fetch these three types of data simultaneously. Specifically, the basis matrix is fetched first and stored stationary within the RE until the associated computations are completed; the weights are then rebuilt in an RE where each row of a coefficient matrix stays stationary until all its associated computations are finished. The third path of MUX1 {\color{black}{\ding{184}}} for the original weights is to handle DNNs' layers where \textit{SmartDeal} is not applied on.

\ul{Fourth, the handling of compact models:} when handling compact models, we consider an adjusted dataflow and PE line configuration for improving the utilization of both the PE slice array and the MAC array within each PE line. Specifically, for depth-wise CONV layers, since the number of CONV channels is only 1, the $dim_C$ PE lines will no longer correspond to input channels. Instead, we map the $R$ number of 1D CONV operations along the dimension of the weight height to these PE lines. 
For squeeze-and-excite/FC layers, each PE line's MAC array of $dim_F$ MACs can be divided into multiple clusters (e.g., two clusters for illustration in the top part of Figure~\ref{fig:overview_PE} (b)) with the help of the two REs in one PE line (denoted as {\color{black}{\ding{185}}} and {\color{black}{\ding{186}}}) and multiplexing units at the bottom of the MAC array, where each cluster handles computations corresponding to a different output pixel in order to improve the MAC array's utilization and thus latency performance. %
In this way, the proposed \textit{SmartDeal} accelerator's advantage is maintained even for compact models, thanks to this adjustment together with 1) our adopted 1D row stationary dataflow within PE lines, 2) the employed bit-serial multipliers, and 3) the possibility to heavily quantized coefficients.

\textbf{Buffer design.}\label{sec:GB} 
For making use of DNNs' (filter-/vector-wise or bit-level) sparsity for skipping corresponding computations/memory-accesses, it in general requires a larger buffer (than that of corresponding dense models) due to the unknown dynamic sparsity patterns. We here discuss how we balance between the skipping convenience and the increased buffer size. Specifically, to enable the processing with sparsity, the row pairs of non-zero input activations and coefficients are selected from the Input GB and the Index GB (using the corresponding coefficient indexes), respectively, as inspired by~\cite{zhou2018cambricon}, which are then sent to the corresponding PE lines for processing with the resulting outputs being collected to the output GB. 

\ul{First, input GB:} to ensure a high utilization of the PE array, a vanilla design requires $(dim_C \times dim_F \times bits_{input})\times$ input activation rows (than that of the dense model counterpart) to be fetched for dealing with the dynamic sparsity patterns, resulting in $(dim_C \times dim_F \times bits_{input})\times$ increased input GB bandwidth requirement. In contrast, our design leads to a $\geq 1/S$ reduction of this required input GB bandwidth, with $dim_C \times dim_F \times bits_{input}$ inputs for every ($S$ + ``Booth encoded non-zero activation bits") cycles. This is because all the FIFOs in the PE lines are implemented in a ping-pong manner using double buffers, thanks to the fact that 1) the adopted 1-D row stationary dataflow at each PE line helps to relieve this bandwidth requirement, because each input activation row can be reused for $S$ cycles; and 2) the bit-serial multipliers takes $\geq1$ cycles to finish an element-wise multiplication.

\ul{Second, weight/index/output buffer:} Similar to that of the input GB, weight/index buffer bandwidth needs to be expanded for handling activation sparsity, of which the expansion is often small thanks to the common observation that the vector-wise activation sparsity ratio is often relatively low. 
Note that because basis matrices need to be fetched and stored into the RE before the fetching of coefficient matrices and the weight reconstruction computation, computation stalls occur if the next basis matrix is fetched after finishing the coefficient fetching and the computation corresponding to the current basis matrix. 
Therefore, we leverage the two REs ({\color{black}{\ding{185}}} and {\color{black}{\ding{186}}} paths) in each PE line to operate in a ``ping-pong" manner to avoid the aforementioned computation stalls. 
For handling the output data, we adopt an FIFO to buffer the outputs from each PE slice before writing them back into the GB, i.e., a cache between the PE array and the output GB. This is to reduce the required output GB bandwidth by making use of the fact that each output is calculated over several clock cycles.

\section{Experiments} \label{sec:results}

In this section, we present a thorough evaluation of \textit{SmartDeal}. We lay out our plan below.

\ul{On the algorithm level}, as \textit{SmartDeal} unifies three mainstream model compression ideas: \textit{sparsification/pruning}, \textit{decomposition}, and \textit{quantization} into one framework, we first present a carefully designed ablation study in Section \ref{sec:exp-inference-ablation} that investigates the effects of different components in \textit{SmartDeal} and the interactions among them.

We then perform extensive experiments (benchmark over two structured pruning and four quantization, i.e., state-of-the-art compression techniques on four standard DNN models with two datasets) to validate its superiority. In addition, we evaluate \textit{SmartDeal} on two compact DNN models (MobileNetV2~\cite{sandler2018mobilenetv2} and EfficientNet-B0~\cite{tan2019efficientnet}) on the ImageNet \cite{Deng09imagenet} dataset, one segmentation model (DeepLabv3+~\cite{chen2017deeplab}) on the CamVid \cite{brostow2008segmentation} dataset, two MLP models on MNIST, and language modeling tasks.

Following the validation of \textit{SmartDeal} inference, we evaluate \textit{SmartDeal} for training on \textit{fine-tuning} and \textit{adaptation} tasks given pre-trained models, which is the common practice in edge-based training \cite{chen2020lottery1,chen2020lottery2}. Extensive training experiments in Section \ref{sec:exp-train-ablation} of two light-weight networks show the storage- and energy-efficiency of \textit{SmartDeal} during training without trading too much model performance. We also provide evaluation over state-of-the-art GPUs in terms of their energy efficiency in Section \ref{sec:exp-train-on-device} by using a state-of-the-art training accelerator \cite{kim20192}.

\ul{On the hardware level}, as the goal of the proposed \textit{SmartDeal} is to boost hardware acceleration energy efficiency and speed, we evaluate \textit{SmartDeal}'s algorithm-hardware co-design results with state-of-the-art DNN accelerators in terms of energy consumption and latency when processing representative DNN models and benchmark datasets. Furthermore, to provide more insights about the proposed \textit{SmartDeal}, we perform various ablation studies to visualize and validate the effectiveness of \textit{SmartDeal}'s component techniques.

\subsection{An Ablation Study on \textit{SmartDeal}'s Building Blocks}
\label{sec:exp-inference-ablation}

\begin{table*}[!t]
    \centering
    \caption{Ablation study of the components in \textit{SmartDeal} using ResNet18 on CIFAR-10 \cite{cifar10} and FPGA energy results. Exp. 7, 8, SD, SD$^\dagger$ use Huffman coding representations for $C_e$ and 8-bit fixed-point representations for $B$ and activations. Other experiments use 32-bit floating-point representations.}
    \begin{tabular}{p{16pt} p{14pt} p{14pt} p{14pt} p{14pt} p{14pt} c c c c c}
    \toprule
    No.          & DE     & Q2     & US     & SS     & RT     & \makecell{Acc}        & \makecell{FLOPs$^*$\\(M)} & \makecell{Size$^\ddagger$\\(MB)} & \makecell{$W$/$C_e$\\Sparsity$^\Box$} & \makecell{Norm.\\E.E.} \\
    \midrule
    1            &        &        &        &        &        & 94.73\%     & 1110.8       & 42.59 & 100.0\%/\hspace{10pt}-\hspace{12pt}  & 1$\times$ \\
    \midrule
    2            & \cmark &        &        &        &        & 94.73\%    & 1177.8       & 43.28 & 100.0\%/100.0\% & 0.79$\times$ \\
    3            & \cmark & \cmark &        &        &        & 94.16\%     & 1177.8       & 11.34 & 100.0\%/100.0\% & 2.49$\times$ \\
    \midrule
    4$^+$            &        &        & \cmark &        &        & 52.98\%     & 212.0        & 9.46  & 19.09\%/\hspace{10pt}-\hspace{12pt} & 2.14$\times$\\
    5            & \cmark &        & \cmark &        &        & 93.36\%     & 689.6        & 9.46  & 37.09\%/20.60\% & 1.71$\times$\\
    6            &        & \cmark & \cmark &        &        & 90.42\%     & 440.5        & 3.97  & 39.66\%/\hspace{10pt}-\hspace{12pt} & 3.77$\times$\\
    7            & \cmark & \cmark & \cmark &        &        & 92.29\%     & 400.0        & 2.95  & 67.22\%/46.44\% & 3.63$\times$\\
    8 & \cmark & \cmark &        &  \cmark                 &  & 43.96\%     & 290.9        & 2.43  & 41.47\%/29.19\% & 3.97$\times$\\
    8$_\vartriangle$ & \cmark & \cmark &  & \cmark &          & 11.30\%     & 233.9        & 1.91  & 20.53\%/13.76\% & 4.26$\times$\\
    \midrule
    9$^+$        &        & \cmark & \cmark &        & \cmark & 93.02\%     & 141.7        & 2.18  & 12.75\%/\hspace{10pt}-\hspace{12pt} & 4.31$\times$\\
    SD           & \cmark & \cmark & \cmark &        & \cmark & 94.32\%     & 261.2        & 1.78  & 22.12\%/12.77\% & 4.15$\times$\\
    SD$^-$       & \cmark & \cmark & \cmark &        & \cmark & 94.60\%     & 552.2        & 2.99  & 48.27\%/36.36\% & -           \\
    SD$^{\dagger}$ & \cmark & \cmark &  & \cmark & \cmark     & 93.30\%     & 209.8        & 1.74  & 17.16\%/10.53\% & 4.38$\times$\\
    SD$^{\dagger}_\vartriangle$ & \cmark & \cmark &  & \cmark & \cmark & 92.54\% &   189.6 & 1.67  & 13.14\%/\ \ 7.75\% & 4.40$\times$\\
    \midrule
    \multicolumn{6}{c}{\textbf{Improv. (SD$^\dagger$ vs. baseline 1)}} & \textbf{-1.43\%} & \textbf{-81.11\%} & \textbf{-95.91\%} &  \textbf{-82.84\%}/\textbf{-89.47\%} &  \textbf{4.38$\times$} \\
    \bottomrule
    \end{tabular}
    \begin{minipage}{0.77\textwidth}
        \begin{tablenotes}
            \item{$\dagger$ Full form of \textit{SmartDeal} with structured sparsity.}
            \item{$\ddagger$ Representations of $C_e$ and $B$ and encoding overheads are all considered for \textit{SmartDeal} models.}
            \item{+ The model is pruned to be of the same model size compared with its counterpart.}
            \item{$-$ A smaller threshold $\theta$ is used in \textit{SmartDeal} for performance comparable to the baseline in experiment No. 1.}
            \item{$\Box$ We provide the sparsity of $C_e$ coefficients as long as DE is involved, in which case we also report the sparsity of rebuilt weight matrices. Sparsity refers to the ratio of \textbf{non-zeros}.}
            \item{$\vartriangle$ Use larger layer-wise structural pruning ratios. See Tab.     \ref{table:ss-ratio-resnet18}.}
        \end{tablenotes}
    \end{minipage}
    \label{table:ablation-inference}
\end{table*}

We first investigate the influence of different components of the \textit{SmartDeal} algorithm:
1) \textit{Decomposition} -- decomposing each layer $W$ into $B$ and $C_e$\, with \textit{no constraint} on $C_e$;
2) \textit{Quantization} -- requiring $C_e$ elements to either power-of-2 values or zero;
3) \textit{Unstructured Sparsification} -- promoting element-wise sparsity in $C_e$;
4) \textit{Structured Sparsity} -- incorporating structured sparsity for $C_e$; and
5) \textit{Re-Training} -- iteratively performing re-training and \textit{SmartDeal}.
For simplicity, we term the above five components as \textbf{DE} (\textbf{DE}composition), \textbf{Q2} (\textbf{Q}uantization to Power-of-\textbf{2}), \textbf{US} (\textbf{U}nstructured \textbf{S}parsity), \textbf{SS} (\textbf{S}tructured \textbf{S}parsity), and \textbf{RT} (\textbf{R}e-\textbf{T}raining), respectively.

Tab. \ref{table:ablation-inference} summarizes the results of ResNet18 \cite{he2016deep} on CIFAR-10 dataset~\cite{cifar10}, with the accuracy, FLOPs \footnote{We use the method in \cite{Micronet-chanllenge} to compare fixed-point and 32-bit floating-point where the highest bit precision of the operands determines the equivalent floating-point operation. Only CONV and FC layers are considered.}, the model storage (total memory size to save all parameters including index if necessary), sparsity levels, and normalized energy efficiency results measured on an edge FPGA (i.e., Ultra 96 FPGA \cite{ultra96}) reported.
For US, we pick the sparsity threshold $\theta=8\PLH10^{-3}$. For RT, we iteratively perform re-training and \textit{SmartDeal} algorithm (with fixed $\theta$) for 80 epochs and report best performing models among.
For SS, the pruning ratios for different layers are manually tuned. We provide the reproducible details for selecting layer-wise pruning ratios in Appendix~\ref{sec:exp-train-settings}, as well as the general rule how we select them.
We use Huffman coding representations for $C_e$ (refer to Section~\ref{subsec:encoding}) and 8-bit fixed-point representations for $B$ and input/output activations for \textit{SmartDeal} (Exp. 7, 8, SD, and SD$^\dagger$). Other experiments use standard 32-bit floating-point representations for input/activations.
We further use a Ultra96 FPGA \cite{ultra96} platform for real-device energy measurement and report the energy efficiency (normalized to the implementation without any components of the \textit{SmartDeal} algorithm). The difference in utilized computation resource on the FPGA board (i.e., digital signal processing unit (DSP) and look-up table (LUT)) among all implementations is within 5\% so that the influence of computation resources is negligible.

\begin{itemize}
    \item \textbf{Quantization}.
        Comparing Exp. 2 and 3, the power-of-2 quantization has aggressive advantage in reducing the model size, without incurring much accuracy loss: $>$\textit{3.7$\times$ reduction in model size while incurring $<$0.6\% accuracy loss}. Note that because we don't quantize $B$ in Exp. 3, the equivalent FLOPs is the same as Exp. 2.
    \item \textbf{(Structured) sparsity}.
        Comparing Exp.~7 with 3, sparsity trades slightly more accuracy drop for higher storage saving and has an immediate influence on FLOPs. Structured sparsity can bring much more aggressive size/FLOPs advantage but will suffer from significant performance loss as shown in Exp. 8. Fortunately, however, model performance can be recovered by re-training in both cases (see last three rows).
    \item \textbf{Decomposition}.
        The benefit brought by decomposition can be supported by comparing multiple pairs of experiments in different cases. For example, when we compare Exp. 6 and 7 (we compress the models to around the same FLOPs number for fair comparison), the model in Exp. 7 (with decomposition) achieves near 2\% higher accuracy with even lower FLOPs.
    \item \textbf{Comparison with sparsity-only models.}
        Comparing (4$^+$, 5) and (9$^+$, SD) shows that under the same tight storage budget, DE induces a much smaller accuracy drop (93.36\% vs. 52.98\% and 94.32\% vs. 93.02\%) at the cost of more FLOPs, which further supports the important role that DE plays in \textit{SmartDeal} by identifying a higher-order structure. The higher FLOPs of Exp.~5 and SD are mainly due to lower sparsity in reconstructed $W$, which has direct influence on FLOPs. Overheads for rebuilding $W$ only account for $<$1\% of total FLOPs thanks to the high sparsity and readily-quantized-to-2 structure of $C_e$.
    \item \textbf{Energy efficiency on FPGA.}
        \textit{SmartDeal} trades higher-cost memory storage/access for lower-cost computation. Therefore energy efficiency is a better way than FLOPs to show its power. We can see that SD enjoys near 2\% better accuracy than Exp.~9$^+$ and comparable energy efficiency despite of higher FLOPs. When favored with structured sparsity, SD$^\dagger$ still outperforms Exp.~9$^+$ with smaller model size and further improves the energy efficiency.\
    \item \textbf{Controlling pruning ratios for structured sparsity.}
        Exp.~8 from Tab.~\ref{table:ablation-inference} shows that structured sparsity improves energy efficiency but at the cost of a drastic accuracy drop, although re-training can recover most of the accuracy in Exp. SD$^\dagger$. Here, we use the additional Exp. 8$_\vartriangle$ and SD$^\dagger_\vartriangle$, in which we use larger pruning ratios for structured sparsity, to show that we can easily trade-off between storage- and energy-efficiency with accuracy by controlling the pruning ratios.
\end{itemize}

\vspace{-0.8em}
\subsection{Evaluation of the \textit{SmartDeal} Algorithm }
\label{subsec:compression_results}

\begin{figure}[!t]
    \vspace{-1em}
    \centerline{\includegraphics[width=90mm]{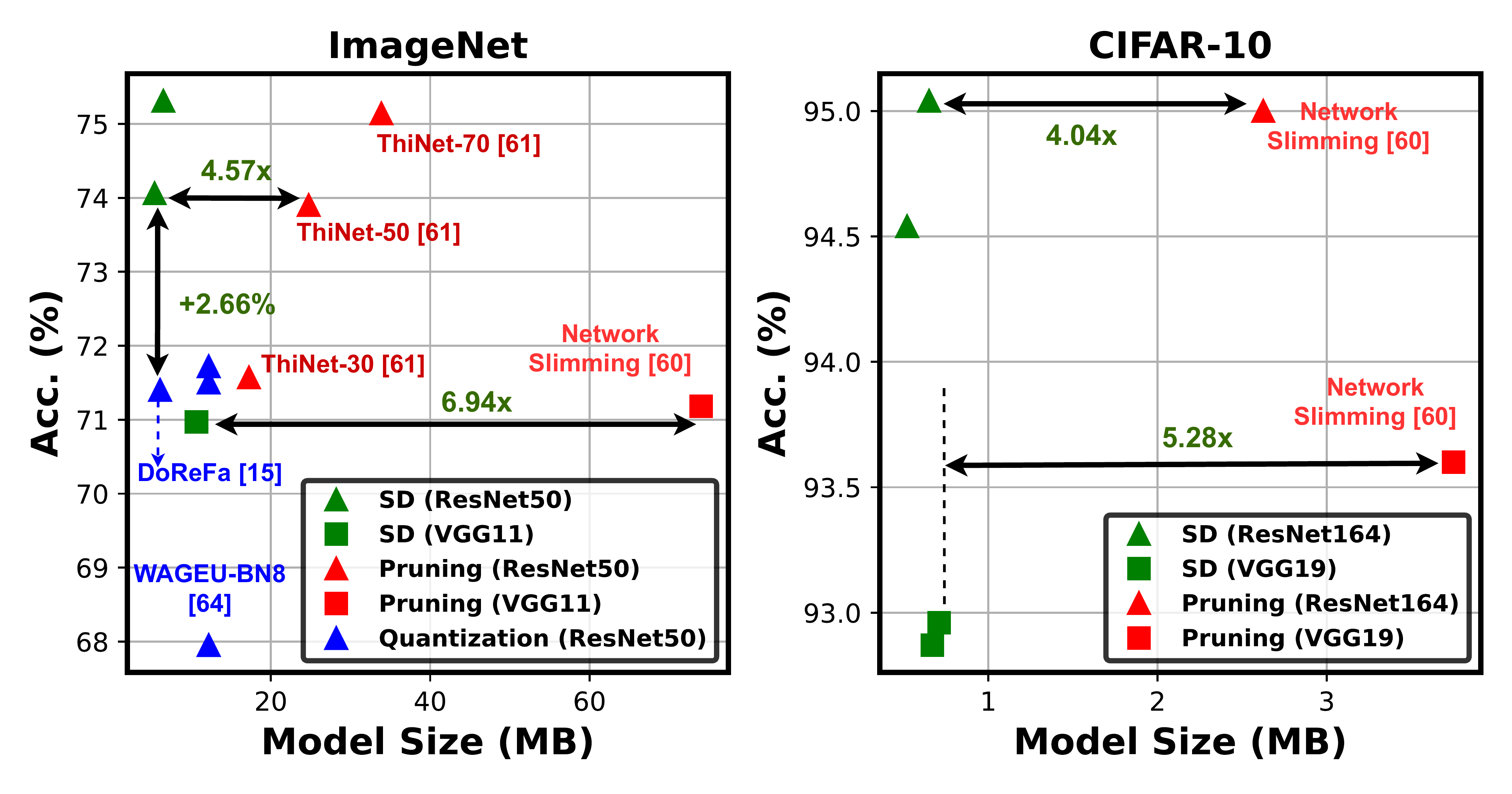}}
        \vspace{-1em}
    \caption{Accuracy vs. model size comparison of the \textit{SmartDeal} algorithm (SE) and state-of-the-art compression techniques on (a) ImageNet and (b) CIFAR-10, where different colors differentiate the SE and baseline techniques. We use 16-bit floating weight representation for the pruning-alone methods.}
    \label{fig:comparison}
        \vspace{-1em}
\end{figure}

\textbf{Experiment settings.}
To evaluate the algorithm performance of \textit{SmartDeal} algorithm, we conduct experiments on 1) a total of \textbf{six DNN models} using both the CIFAR-10 \cite{cifar10} and ImageNet \cite{Deng09imagenet} datasets, 2) \textbf{one segmentation model} on the CamVid \cite{brostow2008segmentation} dataset, and 3) \textbf{two MLP models} on the MNIST dataset and compare the performance with state-of-the-art compression techniques in terms of accuracy and model size, including two structured pruning techniques (Network Slimming \cite{network_slimming} and ThiNet\cite{luo2017thinet}\footnote{We re-implement ThiNet\cite{luo2017thinet} and report the best numbers that we reproduce in the same computation environment for fair comparison, because we obtain better baseline accuracies than the original ThiNet paper. Note that we report model size in megabytes (MB) (with 16-bit weight representation) while \textit{the number of parameters} in million (M) is reported in \cite{luo2017thinet}.}), four quantization techniques (Scalable 8-bit (S8)\cite{banner2018scalable}, FP8 \cite{fp8}, WAGEUBN \cite{WAGEUBN}, and DoReFa \cite{zhou2016dorefa}), one power-of-two quantization technique \cite{qin2019accelerating}, and one pruning and quantization technique \cite{zhou2018cambricon}.

\textbf{\textit{SmartDeal} vs. existing compression techniques.}
As \textit{SmartDeal} unifies the three mainstream ideas of pruning, decomposition and quantization, we evaluate the \textit{SmartDeal} algorithm performance by comparing it with state-of-the-art pruning-alone and quantization-alone algorithms\footnote{we did not include decomposition-alone algorithms since their results are not as competitive and also less popular.}, under four DNN models and two datasets. Note that we use 16-bit floating-point weight representation for the pruning-alone algorithms for more fair comparison.
The experiment results are shown in Figure \ref{fig:comparison}. \textit{SmartDeal} in general outperforms all other pruning-alone or quantization-alone competitors, in terms of the achievable trade-off between the accuracy and the model size. Taking ResNet50 on ImageNet as an example, the quantization algorithm DoReFa\cite{zhou2016dorefa} seems to aggressively shrink the model size yet unfortunately cause a larger accuracy drop; while the pruning algorithm ThiNet\cite{luo2017thinet} maintains competitive accuracy at the cost of larger models. In comparison, \textit{SmartDeal} combines the best of both worlds: it obtains almost as high accuracy as the pruning-only ThiNet\cite{luo2017thinet}, which is 2.66\% higher than the quantized-only DoReFa \cite{zhou2016dorefa}; and on the other hand, it keeps the model as compact as DoReFa\cite{zhou2016dorefa}.
SmartDeal also outperforms more recent and competitive model compression works [a1, a3]. When compared with the state-of-the-art quantization method LSQ \cite{esser2019learned} that uses 3-bit weight representation, SmartDeal costs much less storage for a ResNet50 network than using LSQ (6.51MB vs. 9.61MB) while inducing less accuracy drop (0.82\% vs. 1.10\%) on ImageNet. Moreover, SmartDeal also outperforms CLIP-Q \cite{tung2018clip}, a state-of-the-art joint pruning-quantization method, yielding smaller model size than Clip-q (6.51MB vs. 6.70MB) but much higher top-1 accuracy (75.31\% vs. 73.80\%).
\begin{scriptsize}
\begin{table}[!t]
    \centering
    \caption{The result summary of the proposed \textit{SmartDeal} with re-training on: 1) VGG11 and ResNet50 using the ImageNet dataset \cite{Deng09imagenet}; 2) VGG19 and ResNet164 using the CIFAR-10 dataset \cite{cifar10}; and 3) MLP-1\cite{qin2019accelerating} and MLP-2\cite{zhou2018cambricon} using the MNIST dataset.}
    \small{
    \begin{tabular}{p{40pt} p{23pt} p{23pt} p{12pt} p{20pt} p{12pt} p{12pt} p{18pt}} \toprule
    \multirow{2}{*}{Model} & Top-1 & Top-5 & CR & Param. & $B$  & $C_e$ & Spar.  \\
                           & (\%)  & (\%)  &  ($\times$)  &  (MB)  & (MB) & (MB)  & (\%)  \\
    \midrule 

    VGG11        & 71.18\% & 90.08\% & -             & 845.75 & -    & -     & - \\
    $\mathrm{VGG11_{SD}}$   & 70.97\% & 89.88\% & 79.26 & 10.67  & 1.67 & 7.46 & 86.00 \\ \midrule
    ResNet50      & 76.13\%   & 92.86\%   & -           & 102.40 & -    & -    & - \\
    ResNet50$\mathrm{_{SD}}$ & 75.31\%   & 92.33\%   & 15.73  & 6.51 & 1.40 & 4.40 & 45.00 \\
    ResNet50$\mathrm{_{SD}}$ & 74.06\%   & 91.53\%   & 18.93  & 5.41 & 1.40 & 3.30 & 58.60 \\
    \bottomrule
    VGG19 & 93.66\%   & -  & - & 80.13 & - & - & - \\
    VGG19$\mathrm{_{SD}}$ & 92.96\%   & -   & 112.3  & 0.71 & 0.27 & 0.37 & 92.80 \\
    VGG19$\mathrm{_{SD}}$ & 92.87\%   & -   & 119.6  & 0.67 & 0.27 & 0.33 & 93.70 \\
    \midrule
    ResNet164 & 94.58\%   & -   & - & 6.75 & - & - & - \\
    ResNet164$\mathrm{_{SD}}$ & 95.04\%   & -   & 10.38  & 0.65 & 0.25 & 0.34 & 37.60 \\
    ResNet164$\mathrm{_{SD}}$ & 94.54\%   & -   & 12.87  & 0.52 & 0.25 & 0.21 & 61.00 \\
    \midrule
    MLP-1 & 98.47\%   & -   & - & 14.125 & - & - & - \\
    MLP-$\mathrm{1_{SD}}$ & 97.32\%   & -   & 188.3  & 0.075 & 0.01 & 0.065 & 82.34 \\
    \midrule
    MLP-2 & 98.50\%   & -   & - & 1.07 & - & - & - \\
    MLP-$\mathrm{2_{SD}}$ & 98.11\%   & -   & 66.88  & 0.016 & 0.00 & 0.021 & 93.33 \\
    \bottomrule
    \end{tabular}
    }
    \begin{tablenotes}
    \item{1. The baseline models use 32-bit floating-point representations for the weights and input/output activations, so as to benchmark with the best achievable accuracy results in the literature.}
    \item{2. The proposed \textit{SmartDeal} models use 8-bit fixed-point representations for the input/output activations; the Huffman coding representations for the coefficient matrices; and 8-bit basis matrices, respectively.}
    \end{tablenotes}
    \label{table:sed_results}
    \vspace{-2em}
\end{table}
\end{scriptsize}

Apart from the aforementioned works, we also evaluate the \textit{SmartDeal} algorithm with a state-of-the-art power-of-two quantization algorithm \cite{qin2019accelerating} based on the same MLP model with a precision of 8 bits: while having a significantly higher compression rate of 188.3$\times$ (vs. 128$\times$ in \cite{qin2019accelerating}), \textit{SmartDeal} achieves a comparable accuracy (97.32\% vs. 97.35\%), even if \textit{SmartDeal} is not specifically dedicated for FC layers while the power-of-two quantization \cite{qin2019accelerating} does. In addition, compared with the pruned and quantized MLP model in \cite{zhou2018cambricon}, \textit{SmartDeal} achieves a higher compression rate of 66.88$\times$ (vs. 40$\times$ in \cite{zhou2018cambricon}) with a comparable accuracy (98.11\% vs. 98.42\%). 

A more extensive set of evaluation results are summarized in Table~\ref{table:sed_results}, in order to show the maximally achievable gains (and the incurring accuracy losses) by applying \textit{SmartDeal} over the original uncompressed models. In Table~\ref{table:sed_results}, ``CR'' means the \textit{compression rate} in terms of the overall parameter size; ``Param.'', ``$B$'', and ``$C_e$'' denote the total size of the model parameters, the basis matrices, and the coefficient matrices, respectively; ``Spar.'' denotes the ratio of the pruned and total parameters (the higher the better). Without too much surprise, \textit{SmartDeal} compresses the VGG networks by 80$\times$ to 120$\times$, all with negligible (less than 1\%) top-1 accuracy losses. For ResNets, \textit{SmartDeal} is still able to achieve a solid $>$10$\times$ compression ratio. For example, when compressing ResNet50, we find \textit{SmartDeal} to incur almost no accuracy drop, when compressing the model size by 15$\times$ to 18$ \times$.

\textbf{\textit{SmartDeal} applied on compact models. } Table~\ref{table:sed_results} seems to suggest that (naturally) applying \textit{SmartDeal} to more redundant models will have more gains. We thus validate whether the proposed \textit{SmartDeal} algorithm remains to be beneficial, when adopted for well-known compact models, i.e., MobileNetV2 (MBV2) \cite{sandler2018mobilenetv2} and EfficientNet-B0 (Eff-B0) \cite{tan2019efficientnet}. 

As Table \ref{table:compact_results} indicates, despite the original light-weight design, \textit{SmartDeal} still yields promising gains. For example, when compressing MBV2 for $7.69\times$ CR, \textit{SmartDeal} only incurs $\sim$2\% top-1 accuracy and 1\% top-5 accuracy losses. This result is impressive and highly competitive when placed in the context: for example, the latest work \cite{gong2019differentiable} reports $8\times$ compression (4-bit quantization) of MobileNetV2, yet with a 7.07\% top-1 accuracy loss. 

\begin{scriptsize}
\begin{table}[!t]
    \centering
    \caption{Evaluation of \textit{SmartDeal} with re-training on two compact models with the ImageNet dataset \cite{Deng09imagenet}.}
    \small{
    \begin{tabular}{p{40pt} p{23pt} p{23pt} p{12pt} p{20pt} p{12pt} p{12pt} p{18pt}} \toprule
    \multirow{2}{*}{Model} & Top-1 & Top-5 & CR & Param. & $B$  & $C_e$ & Spar.  \\
                           & (\%)  & (\%)  &  ($\times$)  &  (MB)  & (MB) & (MB)  & (\%)  \\
    \midrule 
    MBV2 & 72.19\%  & 90.53\%  & - & 13.92 & - & - & - \\
    MBV2$\mathrm{_{SD}}$ & 70.16\%  & 89.54\%  & 7.69 & 1.81 & 0.37 & 1.44 & 0.00 \\
    \midrule
    Eff-B0 & 76.30\% & 93.50\%   & - & 20.40 & - & - & - \\
    Eff-B0$\mathrm{_{SD}}$ & 73.80\%  & 91.79\%  & 7.82 & 2.61 & 0.51 & 2.10 & 0.00 \\
        \bottomrule
    \end{tabular}
    }
    \label{table:compact_results}
    \vspace{-1em}
\end{table}
\end{scriptsize}

\textbf{Extending \textit{SmartDeal} beyond classification.} While model compression methods (and hence co-design works) are dominantly evaluated on classification benchmarks, we demonstrate that the effectiveness of \textit{SmartDeal} is beyond one specific task setting. We choose semantic segmentation, a popular computer vision task that is well known to be memory/latency/energy-demanding, to apply the proposed algorithm. Specifically, we choose the state-of-the-art DeepLabv3+ \cite{chen2017deeplab} with a ResNet50 backbone (output stride: 16), and the CamVid\cite{brostow2008segmentation} dataset using its standard split. Compared to the original DeepLabv3+, applying \textit{SmartDeal} can lead to 10.86$\times$ CR, with a marginal mean Intersection over Union (mIoU) drop from 74.20\% to 71.20\% (on the validation split). 

\textbf{Extending \textit{SmartDeal} beyond computer vision tasks.} We additionally perform character-level language modeling experiments with Penn Treebank dataset \cite{marcus1993building}. Under the same compression ratio, \textit{SmartDeal} achieves 0.22 better bit-per-character (1.38 vs. 1.60) over standard unstructured pruning.

\subsection{\textit{SmartDeal} Training Evaluation}
\label{sec:exp-train}

\subsubsection{Fine-Tuning and Adaptation Study}\label{sec:exp-train-ablation}

We present experiment results of SD-T in fine-tuning and adaptation tasks to support the efficacy of \textit{SmartDeal} for resource-constrained on-device training. We run training experiments in two settings on partitioned CIFAR-10/100 datasets. Our experiments are based on ResNet18 \cite{he2016deep} and MobileNetV2 \cite{sandler2018mobilenetv2}. We compare SD-T (with and without structured sparsity) with standard training. We denote SD-T as \textbf{SD} and SD-T with structured sparsity as \textbf{SS}. Besides accuracies, we also compare the model sizes and normalized energy efficiency.

We consider two datasets split strategies on CIFAR-10/100, corresponding to two scenarios, \textit{fine-tuning} and \textit{adaptation}:
\begin{itemize}
    \item \textbf{Fine-tuning.}
        Training samples of all classes are split evenly into two non-overlapping subsets (denoted as sets $\alpha$ and $\beta$). Either $\alpha$ and $\beta$ contains all classes. During training, we first pre-train the model with the set $\alpha$ (not counting into the training energy cost comparison). Then, we fine-tune this same pre-trained model using the different above methods over the set $\beta$, expecting to continuously grow the performance. We use the original testing set with no split.
    \item \textbf{Adaptation.}
        The training set is split \textit{by classes} into two non-overlapping subsets, A and B.     We give a 50-50 splitting on CIFAR-10 and a 90-10 splitting on CIFAR-100. During training, we first pre-train the model with the set A (not counting into the training energy cost comparison). Then, starting from this same pre-trained model (except the classification layer being reset to random initialization), we now train it on the set B, using the different above methods, showing how accurately/efficiently the pre-trained model can be adapted to an unseen classification task. For the testing set, we only use the same five class as in B. On CIFAR-100, we adapt from 90 classes the remaining 10.
\end{itemize}

\begin{table}[t]
    \centering
    \small
    \caption{Result of ResNet18 and MobileNet-V2 experiments for fine-tuning and adaptation tasks on CIFAR-10.}
    \begin{tabular}{@{}c c c c c@{}} \toprule
    Task  & & Acc & \makecell{Norm. Energy\\Efficiency} & \makecell{Size(MB)}\\
    \midrule\midrule
     & \multicolumn{4}{c}{ResNet18} \\ \cmidrule(l){2-5}
    \multirow{3}{*}{\makecell{Fine\\Tuning}}
      & VA & 94.67\% & 1    & 42.59 \\
      & SD & 93.62\% & 1.36 & 2.85 \\
      & SS & 92.40\% & 1.50 & 2.44 \\
      \midrule
    \multirow{3}{*}{\makecell{Adapt-\\ation}}
      & VA & 95.80\% & 1    & 42.59 \\
      & SD & 93.60\% & 1.13 & 2.52 \\
      & SS & 93.32\% & 1.34 & 2.34 \\
    \midrule\midrule
     & \multicolumn{4}{c}{MobileNet-V2} \\ \cmidrule(l){2-5}
    \multirow{3}{*}{\makecell{Fine\\Tuning}}
      & VA & 91.26\% & 1    & 8.63 \\
      & SD & 90.98\% & 1.29 & 1.03 \\
      & SS & 90.38\% & 1.40 & 0.99 \\
      \midrule
    \multirow{3}{*}{\makecell{Adapt-\\ation}}
      & VA & 93.66\% & 1    & 8.64 \\
      & SD & 92.52\% & 1.31 & 1.00 \\
      & SS & 92.07\% & 1.43 & 0.98 \\
    \bottomrule
    \end{tabular}
    \label{table:sd-t-c10}
    \vspace{-1.4em}
\end{table}

\begin{figure}[b!]
    \centering
    \vspace{-1em}
    \includegraphics[width=0.4\textwidth]{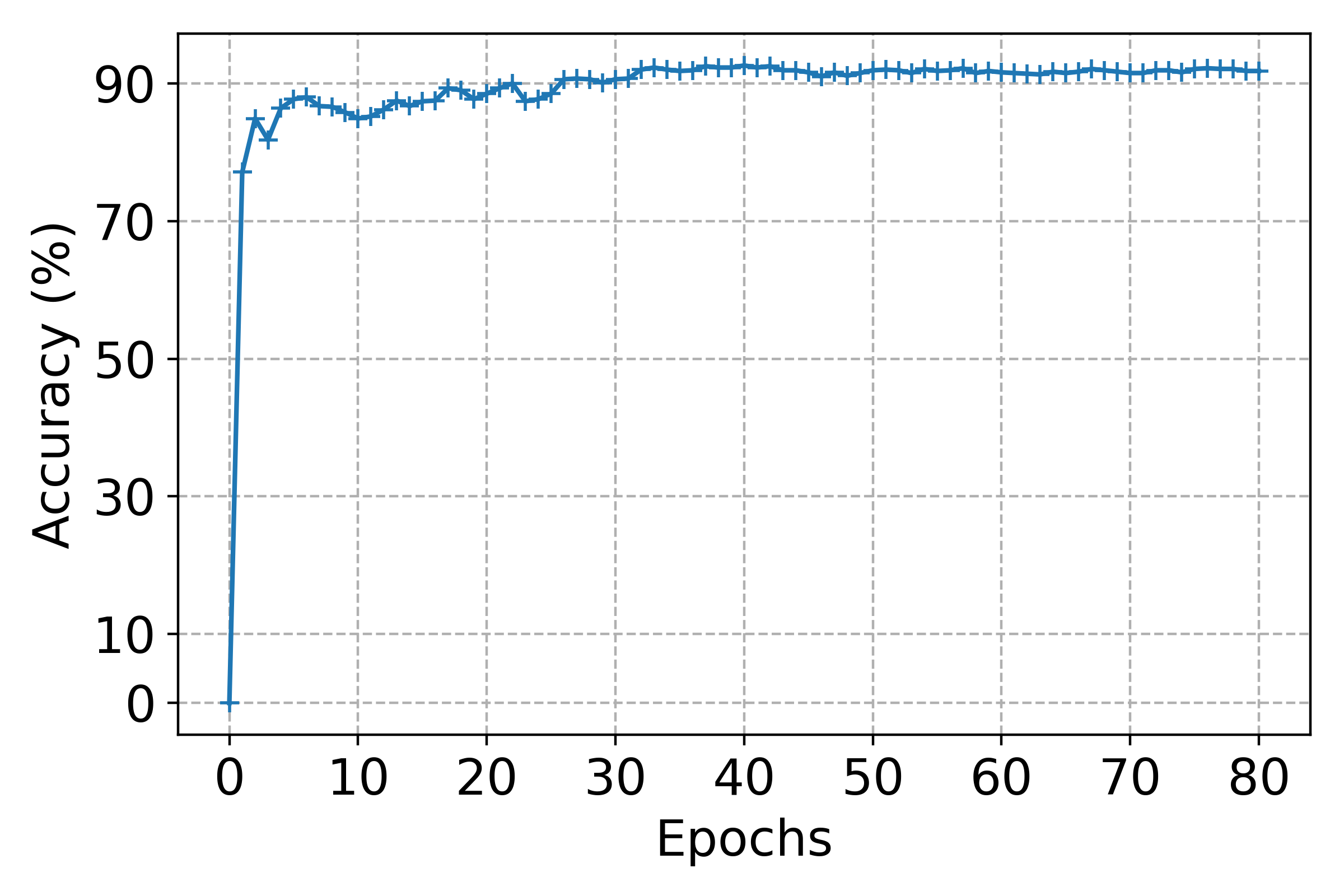}
    \vspace{-1em}
    \caption{Convergence curve of adaptation experiment of SD-T using ResNet-18 on CIFAR-100 dataset.}
    \label{fig:ep80_curve}
\end{figure}

Results of fine-tuning and adaptation experiments of ResNet18 and MobileNetV2 on CIFAR-10/100 datasets are shown in Tab. \ref{table:sd-t-c10} and \ref{table:sd-t-c100}, respectively. Refer to the Appendix~\ref{sec:exp-train-settings} for detailed experiment settings. We evaluate the hardware quantified benefits of \textit{SmartDeal} for training using a state-of-the-art accelerator \cite{kim20192} using the method in \cite{yang2018dnn}. We can see SD-T saves energy during fine-tuning/adaption for both networks. When combined with structured sparsity, SD-T can further save the energy cost with tolerable accuracy drop.
In Figure~\ref{fig:ep80_curve}, we take the adaptation task for ResNet-18 on CIFAR-100 dataset as an example and show the convergence curve of SD-T in terms of testing accuracy on the target classes. We can see that SD-T converges quickly in the first epoch, reaching an accuracy of 77.10\%, and then gradually improves the accuracy to 91.40\% during the remaining training process.
Note that the testing accuracy starts from zero before training starts because we
reset the ﬁnal linear layer for the adaptation task.

\subsubsection{Hardware quantified benefits of \textit{SmartDeal} on-device training}
\label{sec:exp-train-on-device}

In this set of experiments, we deploy SD-T on a state-of-the-art training accelerator \cite{kim20192} and compare the accelerator's performance over SOTA GPUs in terms of energy efficiency. 
Here, we use the state-of-the-art training accelerator \cite{kim20192} instead of designing a dedicated one in order to demonstrate the generality of SD-T algorithm which can potentially improve the performance of training process regardless of the deployment hardware design. 
We follow the evaluation method in \cite{yang2018dnn} to evaluate the accelerator by implementing a cycle-accurate simulator, aiming to model the Register-Transfer-Level (RTL) behavior of the hardware circuits \cite{kim20192}.
As shown in Tab. \ref{table:comparison_training}, by running SD-T-based ResNet18, \cite{kim20192} achieves 1.3$\times$ and 4.48$\times$ improvement in energy efficiency (E.E.) over powerful GPUs with vanilla ResNet18, which are designed in more advanced technologies. 

\begin{table}[!t]
    \centering
    \small
    \caption{Result of ResNet18 and MobileNet-V2 experiments for fine-tuning and adaptation tasks on CIFAR-100.}
    \begin{tabular}{@{}c c c c c@{}} \toprule
    Task  & & Acc & \makecell{Norm. Energy\\Efficiency} & \makecell{Size(MB)}\\
    \midrule\midrule
     & \multicolumn{4}{c}{ResNet18} \\ \cmidrule(l){2-5}
    \multirow{3}{*}{\makecell{Fine\\Tuning}}
      & VA & 73.87\% & 1    & 42.76 \\
      & SD & 72.63\% & 1.34 & 3.92 \\
      & SS & 71.40\% & 1.73 & 3.34 \\
      \midrule
    \multirow{3}{*}{\makecell{Adapt-\\ation}}
      & VA & 92.40\% & 1    & 42.76 \\
      & SD & 91.40\% & 1.50 & 4.04 \\
      & SS & 91.90\% & 1.92 & 3.49 \\
    \midrule\midrule
     & \multicolumn{4}{c}{MobileNet-V2} \\ \cmidrule(l){2-5}
    \multirow{3}{*}{\makecell{Fine\\Tuning}}
      & VA & 68.71\% & 1    & 9.07 \\
      & SD & 69.38\% & 1.27 & 1.21 \\
      & SS & 68.75\% & 1.36 & 1.15 \\
      \midrule
    \multirow{3}{*}{\makecell{Adapt-\\ation}}
      & VA & 89.90\% & 1    & 9.07 \\
      & SD & 90.00\% & 1.33 & 1.57 \\
      & SS & 90.60\% & 1.47 & 1.52 \\
    \bottomrule
    \end{tabular}
    \label{table:sd-t-c100}
    \vspace{-1em}
\end{table}

\begin{table}[!t]
\caption{Peak energy efficiency of ASIC training accelerator \cite{kim20192} with SD-T-based ResNet18 vs. SOTA edge \cite{TX2} and cloud GPU \cite{V100} with ResNet18.}
\label{table:comparison_training}
\centering
    \begin{tabular}{lccc} \toprule
    & \makecell{Freq.\\(MHz)} & \makecell{Tech.\\(nm)} & \makecell{E. E.\\(GOP/s/W)} \\
    \midrule
    Edge GPU \cite{TX2} & 1455 & 40 & 120 \\
    Cloud GPU \cite{V100} & 1300 & 22 & 400 \\
    SD-T~\cite{kim20192} & 500 & 65 & 538 \\ 
    \midrule
    \textbf{Improv.} & & & \textbf{1.34}-\textbf{4.48$\times$}\\
    \bottomrule
    \end{tabular}
\end{table}

\subsection{Evaluation of the Dedicated \textit{SmartDeal} Accelerator.}
\label{sec:hw-performance}

In this subsection, we present experiments to evaluate the performance of the dedicated \textit{SmartDeal} accelerator. Specifically, we first introduce the experiment setup and methodology, and then compare \textit{SmartDeal} accelerator with \textbf{four state-of-the-art DNN accelerators} (covering a diverse range of design considerations) on \textbf{seven DNN models} (including four standard DNNs, two compact models, and one segmentation model) in terms of energy consumption and latency when running on \textbf{three benchmark datasets}. 
Details and ablation studies of the accelerator can be referred from Section V of our prior work \cite{zhao2smartexchange}.

\begin{table}[!t]
    \centering
    \caption{The design considerations of the baseline and our accelerators.}
    \begin{tabular}{@{} c c @{}} 
    \toprule
    Accelerator & Design Considerations \\
    \midrule
     \textbf{DianNao~\cite{chen2014diannao}} & Dense models\\
     \textbf{Cambricon-X~\cite{zhang2016cambricon}} & Unstructured weight sparsity \\
     \multirow{2}{*}{\textbf{SCNN~\cite{parashar2017scnn}}} & Unstructured weight sparsity \\
                                                            &  + Activation sparsity \\
     \textbf{Bit-pragmatic~\cite{albericio2017bit}} & Bit-level activation sparsity \\
     \midrule
     \multirow{2}{*}{\textbf{Ours}} & Vector-wise weight sparsity \\
                                    &  + Bit-level and vector-wise activation sparsity \\
    \bottomrule
    \end{tabular}
    \label{table:considerations}
\end{table}

\textbf{Experiment setup and methodology.} 
\ul{Baselines and configurations:} we benchmark the \textit{SmartDeal} accelerator with four state-of-the-art accelerators: DianNao~\cite{chen2014diannao}, SCNN~\cite{parashar2017scnn}, Cambricon-X~\cite{zhang2016cambricon}, and Bit-pragmatic~\cite{albericio2017bit}. These representative accelerators have demonstrated promising acceleration performance, and are designed with a diverse design considerations as summarized in Table~\ref{table:considerations}. Specifically, DianNao~\cite{chen2014diannao} is a classical architecture for DNN inference which is reported to be over 100$\times$ faster and over 20$\times$ more energy efficient than those of CPUs. While DianNao considers dense models, the other three accelerators take advantage of certain kinds of sparsity in DNNs. To ensure fair comparisons, we assign the \textit{SmartDeal} accelerator and baselines with the same computation resources and on-chip SRAM storage in all experiments, as listed in Table~\ref{table:baselines}. For example, the DianNao, SCNN and Cambricon-X accelerators use 1K 8-bit non-bit-serial multipliers and \textit{SmartDeal} and Bit-pragmatic employ an equivalent 8K bit-serial multipliers.

\begin{figure} [!t]
    \centering
    \includegraphics[width=\linewidth]{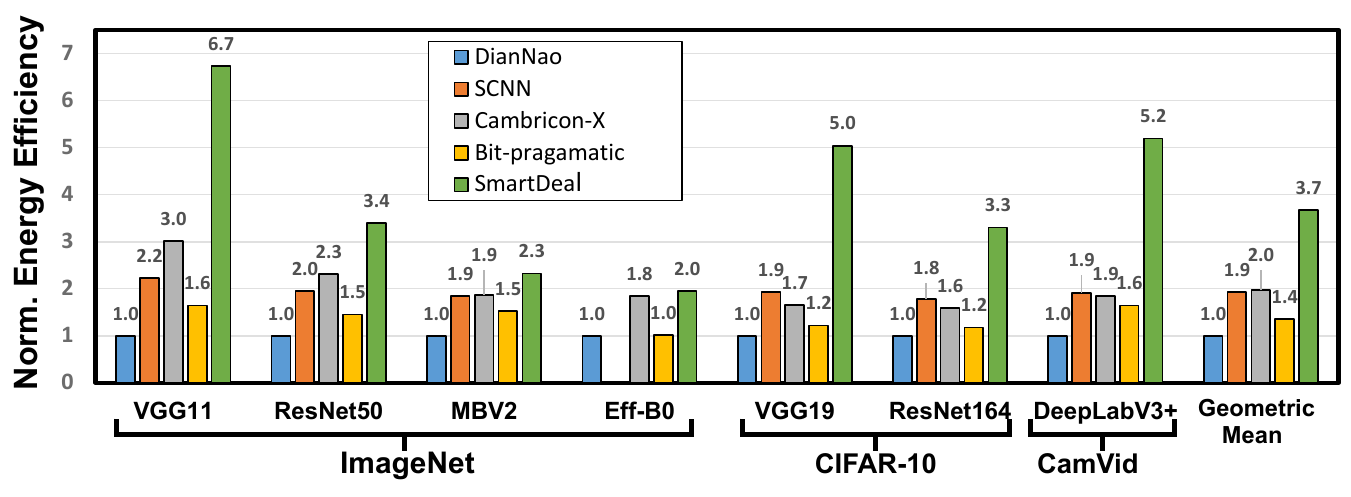}
    \caption{The normalized energy efficiency (over DianNao) achieved by the \textit{SmartDeal} accelerator over the four state-of-the-art baseline accelerators on seven DNN models and three datasets.}
    \label{fig:energy_all}
\end{figure}

\begin{figure}[!t]
    \centering
    \includegraphics[width=\linewidth]{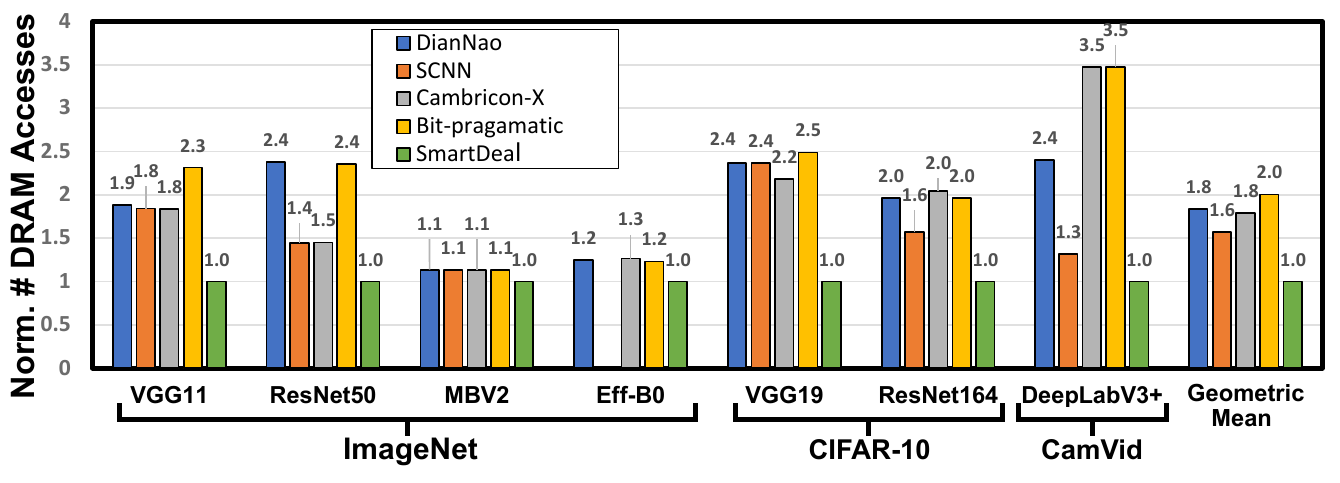}
    \caption{The normalized number of DRAM accesses (over the \textit{SmartDeal} accelerator) of the  \textit{SmartDeal} and four state-of-the-art baseline accelerators on seven DNN models and three datasets.}
    \label{fig:dram}
\end{figure}

\begin{figure}[!t]
    \centering
    \includegraphics[width=\linewidth]{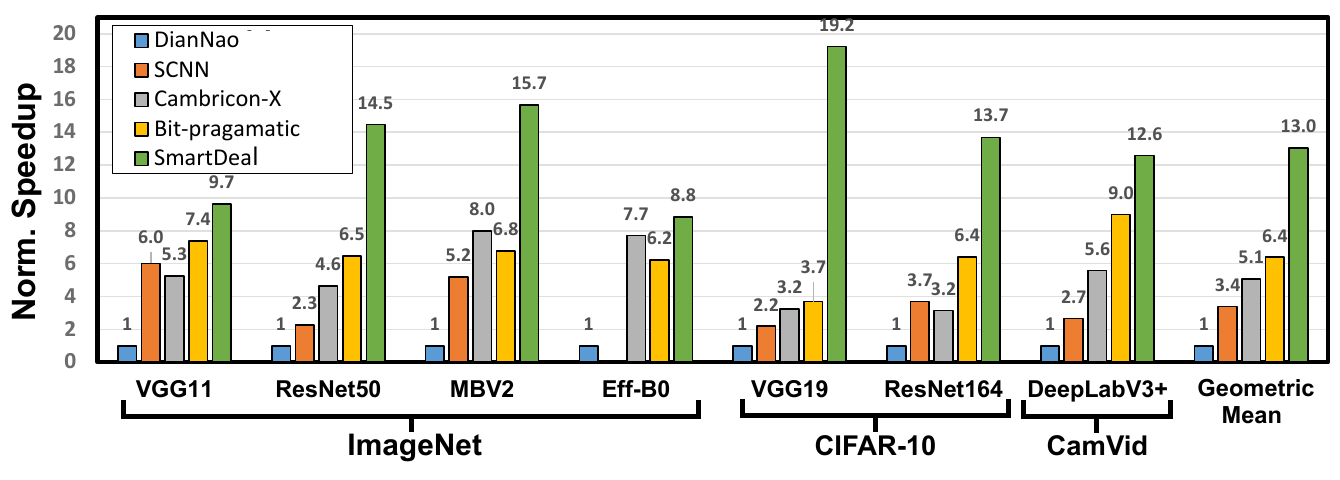}
    \caption{The normalized speedup (over DianNao) achieved by the \textit{SmartDeal} accelerator over the four state-of-the-art baseline accelerators on seven DNN models and three datasets.}
    \label{fig:speedup_all}
\end{figure}

For handling the dynamic sparsity in the \textit{SmartDeal} accelerator, the on-chip input GB bandwidth and weight GB bandwidth with each PE slice are set to be four and two times of those in the corresponding dense models, respectively, which are empirically found to be sufficient for handling all the considered models and datasets. Meanwhile, because the computation resources for the baseline accelerators may be different from their original papers, the bandwidth settings are configured accordingly based on their papers' reported design principles. Note that 1) we do not consider FC layers when benchmarking the \textit{SmartDeal} accelerator with the baseline accelerators (see Figure~\ref{fig:energy_all},\ref{fig:dram},\ref{fig:speedup_all}) for a fair comparison as the SCNN~\cite{parashar2017scnn} baseline is designed for CONV layers, and similarly, we do not consider EfficientNet-B0 for the SCNN accelerator as SCNN is not designed for handling the squeeze-and-excite layers adopted in EfficientNet-B0; 2) our ablation studies consider all layers in the models can be referred from Section V of our prior work \cite{zhao2smartexchange}.

\begin{table}[!t]
    \centering
    \caption{A summary of the computation and storage resources in the \textit{SmartDeal} and baseline accelerators. The area overhead for the RE modules in \textit{SmartDeal} is around 0.64\%.}
    \begin{tabular}{@{}r l r l @{}} 
    \toprule
    \multicolumn{4}{c}{\textbf{\textit{SmartDeal} and Bit-pragmatic~\cite{albericio2017bit} }} \\
    \midrule 
     \textbf{$dim_M$} & 64 & \textbf{Input GB} & 16KB$\times$32Banks   \\
     \textbf{$dim_C$} & 16 & \textbf{Output GB} & 2KB$\times$2Banks    \\
     \textbf{$dim_F$} & 8  & \textbf{Weight Buff./slice} &  2KB$\times$2Banks \\
     \textbf{\# of bit-serial mul.} & 8K  & \textbf{Precision} &  8 bits \\
    \midrule 
    \multicolumn{4}{c}{\textbf{DianNao~\cite{chen2014diannao}, SCNN~\cite{parashar2017scnn}, and Cambricon-X~\cite{zhang2016cambricon}  }} \\
    \midrule 
     \multicolumn{4}{l}{The same total on-chip SRAM storage as \textit{SmartDeal}} \\
     \textbf{\# of 8-bit mul.} & 1K  & \textbf{Precision} &  8 bits \\
    \bottomrule
    \end{tabular}
    \label{table:baselines}
\end{table}

\ul{Benchmark models, datasets, and precision:}
We use seven representative DNNs (ResNet50, ResNet164, VGG11, VGG19, MobileNetV2, EfficientNet-B0, and DeepLabV3+) and three benchmark datasets (CIFAR-10~\cite{cifar10}, ImageNet~\cite{Deng09imagenet}, and CamVid~\cite{brostow2008segmentation}). Regarding the precision, we adopt 1) 8-bit activations for both the baseline-used and \textit{SmartDeal}-based DNNs; and 2) 8-bit weights in the baseline-used DNNs, and 8-bit precision for the basis matrices and Huffman coding for the coefficient matrices in the \textit{SmartDeal}-based DNNs.

\ul{Technology-dependent parameters:} 
For evaluating the performance of the \textit{SmartDeal} accelerator, we implemented a custom cycle-accurate simulator, aiming to model the RTL behavior of synthesized circuits, and verified the simulator against the corresponding RTL implementation to ensure its correctness. Specifically, 
the gate-level netlist and SRAM are generated based on a commercial 28nm technology using the Synopsys Design Compiler and Arm Artisan Memory Compilers, proper activity factors are set at the input ports of the memory/computation units, and the energy is calculated using a state-of-the-art tool PrimeTime PX~\cite{PTPX}. 
Meanwhile, thanks to the clear description of the baseline accelerators' papers and easy representation of their works, we followed their designs and implemented custom cycle-accurate simulators for all the baselines. In this way, we can evaluate the performance of both the baseline and our accelerators based on the same commercial 28nm technology. 
The resulting designs operate at a frequency of 1GHz and the performance results are normalized over that of the DianNao accelerator, where the DianNao design is modified to ensure that all accelerators have the same hardware resources (see Table \ref{table:baselines}). 
We refer to ~\cite{yang2018dnn} for the unit energy of DRAM accesses, which is 100pJ per 8 bit, and the unit energy costs for computation and SRAM accesses are listed in Table \ref{table:unit_energy}.

\textbf{\textit{SmartDeal} vs. state-of-the-art accelerators.} 
\ul{Energy efficiency over that of the baseline accelerators:}
Figure~\ref{fig:energy_all} shows the normalized energy efficiency of the \textit{SmartDeal} and the baseline accelerators. It is shown that the \textit{SmartDeal} accelerator consumes the least energy under all the considered DNN models and datasets, achieving an energy efficiency improvement ranging from $2.0\times$ to $6.7\times$. The \textit{SmartDeal} accelerator's outstanding energy efficiency performance is a result of \textit{SmartDeal}'s algorithm-hardware co-design effort to effectively trade the much higher-cost memory storage/accesses for the lower-cost computations (i.e., rebuilding the weights using the basis and coefficient matrices at the least costly RF and PE levels vs. fetching them from the DRAM). Note that \textit{SmartDeal} non-trivially outperforms all baseline accelerators even on the compact models (i.e., MobileNetV2 and EfficientNet-B0) thanks to both the \textit{SmartDeal} algorithm's higher compression ratio and the \textit{SmartDeal} accelerator's dedicated and effective design (see Section \ref{subsec:accelerator}) of handling depth-wise CONV and squeeze-and-excite layers that are commonly adopted in compact models.

Figure~\ref{fig:dram} shows the normalized number of DRAM accesses for the weights and input/output activations. We can see that: 
1) the baselines always require more (1.1$\times$ to 3.5$\times$) DRAM accesses than the \textit{SmartDeal} accelerator, e.g.,  see the ResNet and VGG models on the ImageNet and CIFAR-10 datasets as well as the segmentation model DeepLabV3+ on the CamVid dataset;
2) \textit{SmartDeal}’s DRAM-access reduction is smaller when the models' activations dominate the cost (e.g., compact DNN models); and
3) the \textit{SmartDeal} accelerator can reduce the number of DRAM accesses over the baselines by up to 1.3$\times$ for EfficientNet-B0, indicating the effectiveness of our dedicated design for handling the squeeze-and-excite layers (see Section \ref{subsec:accelerator}).

\ul{Speedup over that of the baseline accelerators:} Similar to benchmarking the \textit{SmartDeal} accelerator's energy efficiency, we compare its latency of processing one image (i.e., batch size is 1) over that of the baseline accelerators on various DNN models and datasets, as shown in Figure~\ref{fig:speedup_all}. We can see that the \textit{SmartDeal} accelerator achieves the best performance under all the considered DNN models and datasets, achieving a latency improvement ranging from $8.8\times$ to $19.2\times$. Again, this experiment validates the effectiveness of
\textit{SmartDeal}'s algorithm-hardware co-design effort to reduce the latency on fetching both the weights and the activations from the memories to the computation resources. Since the \textit{SmartDeal} accelerator takes advantage of both the weights' \textit{vector-wise} sparsity and the activations' \textit{bit-level} and \textit{vector-wise} sparsity, it has a higher speedup over all the baselines that make use of only one kind of sparsity. Specifically, the \textit{SmartDeal} accelerator has an average latency improvement of 3.8$\times$, 2.5$\times$, and 2.0$\times$ over SCNN~\cite{parashar2017scnn} and Cambricon-X~\cite{zhang2016cambricon} which consider unstructured sparsity, and Bit-pragmatic~\cite{albericio2017bit} which considers the \textit{bit-level} sparsity in activations, respectively.

\section{Conclusion and future work}

The trade-off between the model performance and its speed/computational cost pervasively exists in the research community and industry. In this paper, we propose a more smart trade-off strategy, \textit{SmartDeal}, an algorithm-hardware co-design framework with a unique goal to trade higher-cost memory storage/access for lower-cost computation, in order to achieve storage- and energy-efficient DNN inference and training. 
Extensive experiments including both algorithm and hardware aspects show that \textit{SmartDeal} can effectively resolve the practical bottleneck of high-cost memory storage/access and can aggressively trim down energy cost and model size, while incurring minimal accuracy drops, for both inference and training. 
Moreover, \textit{SmartDeal} provides a handy way for the users to freely balance such trade-off by controlling hyperparameters in \textit{SmartDeal} to favor either better performance or higher speed.
Our immediate future goal is to automate the hyper parameter tuning in \textit{SmartDeal} via AutoML.

\section*{Acknowledgment}

The work is supported by the National Science Foundation (NSF) through the Real-Time Machine Learning program (Award number: 1937592, 1937588).

\ifCLASSOPTIONcaptionsoff
  \newpage
\fi



\bibliographystyle{IEEEtran}
\bibliography{ref}

\clearpage

\appendices

\section{\textit{SmartDeal} Decomposition Evolution. }
\label{sec:evolution}

\begin{figure}[!b]
    \centerline{\includegraphics[width=70mm]{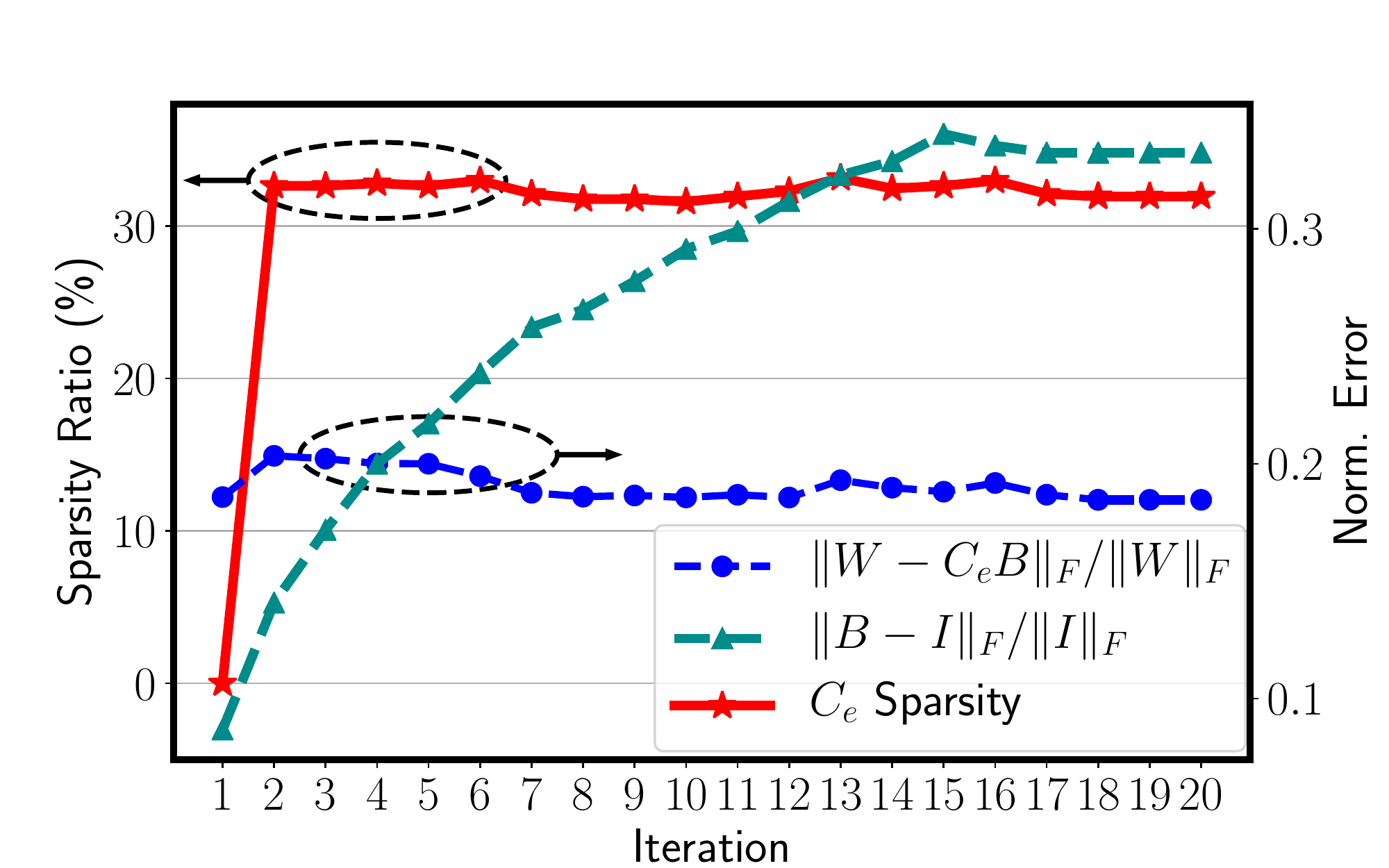}}
    \caption{{Illustrating an example of the solution evolution during the \textit{SmartDeal} algorithm training.}}
    \label{fig:evolution}
\end{figure}

To give an example of the decomposition evolution of the \textit{SmartDeal} algorithm, we take one weight matrix $W\in R^{192\times 3}$ from the second CONV layer of the second block in a ResNet164 network pre-trained on CIFAR-10. The \textit{SmartDeal} algorithm decomposes $W=C_e B$, where $C_e\in R^{192\times 3}$ and $B\in R^{3\times 3}$. Figure \ref{fig:evolution} shows the evolution of the reconstruction error, sparsity ratio in $C_e$, and the distance between $B$ and its initialization (identity). We can see that the sparsity ratio in $C_e$ will increase at the beginning at the cost of an increased reconstruction error. But the \textit{SmartDeal} algorithm remedies the error over iterations while maintaining the sparsity. Also, $B$ will gradually become more different from the initialization.

\section{Encoding $C_e$ and $B$ in \textit{SmartDeal}}
\label{subsec:encoding}

After we obtain $C_e$ and $B$ via \textit{SmartDeal} algorithm, we introduce their storage-economic and hardware-friendly representations. We use the representations explained below for all \textit{SmartDeal} models throughout this paper.

\paragraph{Encoding $B$} Note that $B$ is a small dense matrix without special structures. We use 8-bit fixed-point representations for $B$ because (1) the low bit-width further reduces the model size and (2) the fixed-point representation can fully exploit the benefits of the bit-shifting operations due to power-of-2 coefficients in $C_e$ and thus minimize the overheads when rebuilding the weights from $C_e$ and $B$. The quantization on $B$ can be easily implemented by adding a quantization step after each fitting step of $B$ in Algorithm~\ref{SD}.

\paragraph{Encoding $C_e$}
When encoding $C_e$, an efficient indexing method is used to store the non-zero coefficient in a compact way as in \cite{zhou2018cambricon}. The ``indices'' are coded with 1-bit: ``0'' for zero coefficients and ``1'' for non-zero coefficients. By using such index, we merely have to store the non-zero coefficients.
The cost of the indexing part can be further minimized by exploiting the structured sparsity in $C_e$ when it is enforced.

\begin{figure}[!t]
    \centerline{\includegraphics[width=70mm]{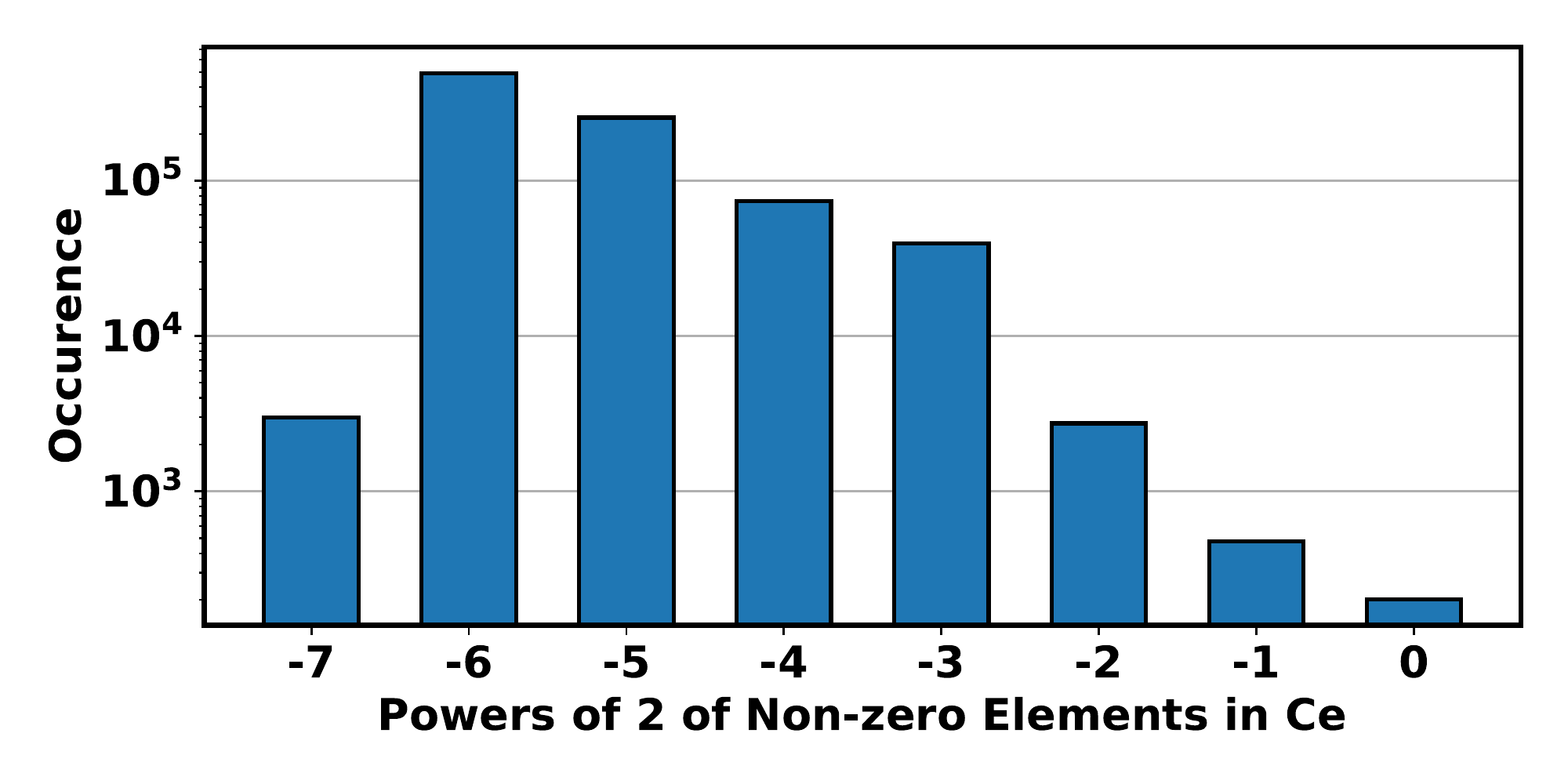}}
    \caption{Highly non-uniform distributions of the powers with base 2 of the non-zero elements in $C_e$ within a \textit{SmartDeal} ResNet18 model trained on the CIFAR-10 dataset.}
    \label{fig:ce_dist}
\end{figure}

For the non-zero elements in $C_e$, we apply Huffman coding representation following \cite{han2015deep}, as we empirically observe highly non-uniform distributions of the non-zero elements. We take from the experiments in Section~\ref{sec:exp-inference-ablation} a ResNet18 model with \textit{SmartDeal} applied (the ``SD'' row in Table~\ref{table:ablation-inference}) and plot the distribution of the powers with base 2 of the non-zero elements in all $C_e$ matrices in Figure~\ref{fig:ce_dist}. In this case, we only need $\sim$2.3 bits on average to represent the 16 possible values (we have 8 options for the powers and two for the signs of the elements). Based on our empirical observations, the effective bit-width for $C_e$ ranges from 2 to 3.5, depending on the model types and tasks.

\section{More Discussions on the \textit{SmartDeal} Rationale}
\label{subsec:sd-discussions}

\paragraph{Rationale of SD} From an \textit{algorithmic} perspective, our SD potentially explores higher-order weight structures, as observed in \cite{Yu_2017_CVPR,gui2019adversarially}. That is, rather than enforcing sparsity, quantization, etc. on the original weight directly, it is often more effective to do so on corresponding decomposed matrix factors (either additive or multiplicative).
From a \textit{hardware} perspective, the resulting $B$ and $C_e$ in SD
enables much reduced memory accesses at a extra light cost of shift-and-add operations.
Considering that the unit energy cost of memory accesses is much higher \cite{eyeriss},
it is promising for energy efficiency and acceleration by trading higher cost memory accesses for lower cost shift-and-add computations. 

\paragraph{Comparison with model compression}
Although the proposed \textit{SmartDeal} algorithm includes popular techniques such as pruning, quantization and matrix decomposition that are intensively investigated and used in model compression, there are major differences between \textit{SmartDeal} and classic model compression works. \textit{SmartDeal} is motivated by the high-cost memory access consumed by DRAM, which is totally neglected by previous model compression works that only pursue high compression ratio in model storage. To achieve this goal, we design a special weight structure that combines pruning and weight decomposition and is DRAM-friendly, and propose \textit{SmartDeal} algorithm to enforce that structure on a trained DNN with minimal performance loss, at the cost of inducing extra computations during inference.
However, this cost can be further decreased by introducing the power-of-2 quantization and reduce expensive floating point multiplications to cheap bit-shifting operations.
In other words, \textit{SmartDeal} is a novel algorithm that aims at ``smartly'' trade for high-cost DRAM access for cheap bit-shifting operations, an opportunity largely overlooked by classic model compression.

\section{Implementation Details for Bucket Switch Updating in Section~\ref{sec:SmartDeal_Training}}
\label{sec:bucket-switch}

During the training process, we utilize ``bucket switch updating'' and ``update decay and aggregation'' for $C_e$'s non-zeros as discussed Section~\ref{sec:SmartDeal_Training}. 
In practice, we will apply a threshold $\theta_g$ to suppress the effect of noisy gradients for ``bucket switch updating'': any gradient whose magnitude is smaller than $\theta_g$ will be treated as zero. Intuitively, a small update on float-point weights will likely be ignored after quantization. We also allow for an ``update delay and aggregation'' scheme, in which we can set a ``counter'' to accumulate the switch decisions in a time window, before actually executing them to update the weight. For example, an element might receive three ``switch up'', one ``switch down'' and one ``unchanged'' pointers, in the last five iterations. while the element is never ``actually'' updated. We can now aggregate those five and record $+ 2$ in the counter. We only switch up or down when the value in the counter reaches $+\theta_c$ or $-\theta_c$, where $\theta_c$ is a positive integer hyper parameter.

\section{Detailed settings of fine-tuning and adaptation experiments in Section~\ref{sec:exp-train}}
\label{sec:exp-train-settings}

\subsection{Training settings}
In both tasks, ResNet18 and MobileNetV2 are pre-trained with an initial learning rate $0.1$. The Stochastic Weight Averaging (SWA) \cite{izmailov2018averaging} will start to be applied at the 140th epoch with a learning rate of 0.05. After pre-training, one pass of \textit{SmartDeal} algorithm (with and without structured sparsity) is applied.

Then, we will train SD-T (with or without structured sparsity) models on the target training subset (B or $\beta$). All networks are trained for 80 epochs and SWA starts at the 40th epoch but we use different initial learning rates and SWA-stage learning rates for different combinations of models and tasks. Settings that work well on CIFAR-10 are found to work well on CIFAR-100 dataset, too.
\begin{itemize}
    \item \textbf{Initial learning rates.} For the fine-tuning task, we set the initial learning rate to $10^{-2}$. For the adaptation task, the networks usually need a more significant change, especially for the linear layers (we reset linear layers of models trained on adaptation tasks). Therefore, we select an initial learning rate of $10^{-1}$.
    \item \textbf{SWA learning rates.} For the fine-tuning task, we use a universal SWA-stage learning rate $2\PLH10^{-3}$ for both models on both datasets. For the adaptation task, ResNet18 uses $5\PLH10^{-3}$ for the SWA-stage learning rate and MobileNetV2 uses $10^{-2}$.
\end{itemize}

\subsection{\textit{SmartDeal} settings}
In the one-pass \textit{SmartDeal} algorithm after pre-training as mentioned above, we use a threshold $\theta=7\PLH10^{-3}$ for both ResNet18 and MobileNetV2.

When we train SD-T models, we select $\theta_c=7$ and $\theta_g=5\PLH10^{-3}$ for ResNet18. The selection of $\theta_g$ for MobileNetV2 is dependent on the task -- we select $5\PLH10^{-3}$ and $10^{-2}$ for the fine-tuning and adaptation tasks, respectively.

\subsection{Data splitting}
For the fine-tuning task, the splitting strategy is the same on CIFAR-10 and CIFAR-100 datasets: 50\% of the data are selected as the \textit{source} or the $\alpha$ set, on which models are pre-trained; the rest are selected as the \textit{target} or the $\beta$ set.

However, it is different for the adaptation task on the two datasets. We use a 50\%-50\% splitting on CIFAR-10 but a 90\%-10\% splitting on CIFAR-100, in which the 90\% part is the \textit{source} set and the remaining 10\% part is the \textit{target}.

\subsection{Layer-wise structured pruning ratios for ResNet18}\label{app:ss-ratio}

We set the layer-wise structural pruning ratios for ResNet18 as in Tab. \ref{table:ss-ratio-resnet18}. The ratio selection follows the following rules: 1) we select smaller ratios in shallow layers; 2) we select larger ratios for larger layers (more redundancy); 3) we select smaller ratios for shortcuts ($1\PLH1$ CONV layers); 4) we simply skip linear layers, since ResNet18 has a very small linear layer that hardly affects performance; 5) we choose smaller pruning ratios on CIFAR-100 dataset than we do on CIFAR-10 dataset.

\subsection{Layer-wise structured pruning ratios for MobileNetV2}\label{app:ss-ratio-mnv2}

We set the layer-wise structural pruning ratios for MobileNetV2 as in Tab. \ref{table:ss-ratio-mobilenetv2}. In addition to the ratio selection principles described in Section \ref{app:ss-ratio}, we also skip the depth-wise separable convolutional layers in MobileNetV2.

\section{Training Accelerator Configuration for \textit{SmartDeal} in Section~\ref{sec:exp-train-on-device}}

The selected state-of-the-art training accelerator \cite{kim20192} can be configured for feed-forward pass, back propagation, and weight gradient calculations with balanced computation and memory bandwidth requirements. We mainly modify this training accelerator for two parts: inserting rebuilding engines for the original weights and inserting counters to store the last five switch directions for bucket switch updating of $C_e$'s non-zero elements. In addition, we reuse the gradients calculations for coefficient and basis matrices.

\begin{table}[t]
    \centering
    \caption{Pruning ratios for different layers in ResNet18 in CIFAR-10 and CIFAR-100 experiments. C$p$ means the $p$th convolutional layer. SC represents shortcut layers. $_\vartriangle$ The larger structural pruning ratios used in experiments marked with a $_\vartriangle$ in Tab. \ref{table:ablation-inference}.}
    \label{table:ss-ratio-resnet18}
    \begin{tabular}{ l  l  c c } \toprule
    Layer & Shape & C-10 & C-100 \\
    \midrule 
    CONV1   & $64  \PLH 3   \PLH 3 \PLH 3$   & 0.2 / 0.4$_\vartriangle$ & 0.2 \\ 
    L1-0-C1 & $64  \PLH 64  \PLH 3 \PLH 3$   & 0.2 / 0.6$_\vartriangle$ & 0.2 \\ 
    L1-0-C2 & $64  \PLH 64  \PLH 3 \PLH 3$   & 0.2 / 0.4$_\vartriangle$ & 0.2 \\ 
    L1-1-C1 & $64  \PLH 64  \PLH 3 \PLH 3$   & 0.2 / 0.4$_\vartriangle$ & 0.2 \\ 
    L1-1-C2 & $64  \PLH 64  \PLH 3 \PLH 3$   & 0.2 / 0.4$_\vartriangle$ & 0.2 \\ 
    L2-0-C1 & $128 \PLH 64  \PLH 3 \PLH 3$   & 0.2 / 0.2$_\vartriangle$ & 0.2 \\ 
    L2-0-C2 & $128 \PLH 128 \PLH 3 \PLH 3$   & 0.3 / 0.3$_\vartriangle$ & 0.2 \\ 
    L2-0-SC & $128 \PLH 64  \PLH 1 \PLH 1$   & 0.1 / 0.1$_\vartriangle$ & 0.1 \\ 
    L2-1-C1 & $128 \PLH 128 \PLH 3 \PLH 3$   & 0.4 / 0.4$_\vartriangle$ & 0.2 \\ 
    L2-1-C2 & $128 \PLH 128 \PLH 3 \PLH 3$   & 0.4 / 0.4$_\vartriangle$ & 0.2 \\ 
    L3-0-C1 & $256 \PLH 128 \PLH 3 \PLH 3$   & 0.3 / 0.3$_\vartriangle$ & 0.2 \\ 
    L3-0-C2 & $256 \PLH 256 \PLH 3 \PLH 3$   & 0.3 / 0.3$_\vartriangle$ & 0.2 \\ 
    L3-0-SC & $256 \PLH 128 \PLH 1 \PLH 1$   & 0.2 / 0.2$_\vartriangle$ & 0.2 \\ 
    L3-1-C1 & $256 \PLH 256 \PLH 3 \PLH 3$   & 0.4 / 0.4$_\vartriangle$ & 0.2 \\ 
    L3-1-C2 & $256 \PLH 256 \PLH 3 \PLH 3$   & 0.5 / 0.6$_\vartriangle$ & 0.2 \\ 
    L4-0-C1 & $512 \PLH 256 \PLH 3 \PLH 3$   & 0.5 / 0.8$_\vartriangle$ & 0.4 \\ 
    L4-0-C2 & $512 \PLH 512 \PLH 3 \PLH 3$   & 0.5 / 0.9$_\vartriangle$ & 0.4 \\ 
    L4-0-SC & $512 \PLH 256 \PLH 1 \PLH 1$   & 0.5 / 0.6$_\vartriangle$ & 0.4 \\ 
    L4-1-C1 & $512 \PLH 512 \PLH 3 \PLH 3$   & 0.5 / 0.9$_\vartriangle$ & 0.4 \\ 
    L4-1-C2 & $512 \PLH 512 \PLH 3 \PLH 3$   & 0.5 / 0.9$_\vartriangle$ & 0.4 \\ 
    LINEAR  & $d_\mathrm{out}  \PLH 512 $    & 0.0                 & 0.0 \\
    \bottomrule
    \end{tabular}
\end{table}

\begin{table}[t]
    \centering
    \caption{
        Pruning ratios for different layers in MobileNetV2 in CIFAR-10 and CIFAR-100 experiments, where C$p$ means the $p$th convolutional layer and SC represents the shortcut layers.
    }
    \label{table:ss-ratio-mobilenetv2}
    \begin{tabular}{ p{40pt}  p{60pt}  c  c } \toprule
    Layer & Shape & C-10 & \makecell{C-100} \\
    \midrule 
    CONV1   &   $ 32 \PLH 3 \PLH 3 \PLH 3 $      &  0.1  &     0.1  \\ 
    L0.C1   &   $ 32 \PLH 32 \PLH 1 \PLH 1 $     &  0.4  &     0.2  \\ 
    L0.C2   &   $ 32 \PLH 1 \PLH 3 \PLH 3 $      &  0    &     0    \\ 
    L0.C3   &   $ 16 \PLH 32 \PLH 1 \PLH 1 $     &  0.8  &     0.4  \\ 
    L0.SC   &   $ 16 \PLH 32 \PLH 1 \PLH 1 $     &  0.2  &     0.2  \\ 
    L1.C1   &   $ 96 \PLH 16 \PLH 1 \PLH 1 $     &  0.2  &     0.2  \\ 
    L1.C2   &   $ 96 \PLH 1 \PLH 3 \PLH 3 $      &  0    &     0    \\ 
    L1.C3   &   $ 24 \PLH 96 \PLH 1 \PLH 1 $     &  0.5  &     0.4  \\ 
    L1.SC   &   $ 24 \PLH 16 \PLH 1 \PLH 1 $     &  0.05 &     0.05 \\ 
    L2.C1   &   $ 144 \PLH 24 \PLH 1 \PLH 1 $    &  0.2  &     0.2  \\ 
    L2.C2   &   $ 144 \PLH 1 \PLH 3 \PLH 3 $     &  0    &     0    \\ 
    L2.C3   &   $ 24 \PLH 144 \PLH 1 \PLH 1 $    &  0.4  &     0.2  \\ 
    L3.C1   &   $ 144 \PLH 24 \PLH 1 \PLH 1 $    &  0.1  &     0.1  \\ 
    L3.C2   &   $ 144 \PLH 1 \PLH 3 \PLH 3 $     &  0    &     0    \\ 
    L3.C3   &   $ 32 \PLH 144 \PLH 1 \PLH 1 $    &  0.1  &     0.1  \\ 
    L4.C1   &   $ 192 \PLH 32 \PLH 1 \PLH 1 $    &  0.2  &     0.2  \\ 
    L4.C2   &   $ 192 \PLH 1 \PLH 3 \PLH 3 $     &  0    &     0    \\ 
    L4.C3   &   $ 32 \PLH 192 \PLH 1 \PLH 1 $    &  0.4  &     0.4  \\ 
    L5.C1   &   $ 192 \PLH 32 \PLH 1 \PLH 1 $    &  0.2  &     0.2  \\ 
    L5.C2   &   $ 192 \PLH 1 \PLH 3 \PLH 3 $     &  0    &     0    \\ 
    L5.C3   &   $ 32 \PLH 192 \PLH 1 \PLH 1 $    &  0.4  &     0.4  \\ 
    L6.C1   &   $ 192 \PLH 32 \PLH 1 \PLH 1 $    &  0.2  &     0.2  \\ 
    L6.C2   &   $ 192 \PLH 1 \PLH 3 \PLH 3 $     &  0    &     0    \\ 
    L6.C3   &   $ 64 \PLH 192 \PLH 1 \PLH 1 $    &  0.4  &     0.4  \\ 
    L7.C1   &   $ 384 \PLH 64 \PLH 1 \PLH 1 $    &  0.2  &     0.2  \\ 
    L7.C2   &   $ 384 \PLH 1 \PLH 3 \PLH 3 $     &  0    &     0    \\ 
    L7.C3   &   $ 64 \PLH 384 \PLH 1 \PLH 1 $    &  0.4  &     0.4  \\ 
    L8.C1   &   $ 384 \PLH 64 \PLH 1 \PLH 1 $    &  0.3  &     0.2  \\ 
    L8.C2   &   $ 384 \PLH 1 \PLH 3 \PLH 3 $     &  0    &     0    \\ 
    L8.C3   &   $ 64 \PLH 384 \PLH 1 \PLH 1 $    &  0.5  &     0.4  \\ 
    L9.C1   &   $ 384 \PLH 64 \PLH 1 \PLH 1 $    &  0.5  &     0.4  \\ 
    L9.C2   &   $ 384 \PLH 1 \PLH 3 \PLH 3 $     &  0    &     0    \\ 
    L9.C3   &   $ 64 \PLH 384 \PLH 1 \PLH 1 $    &  0.6  &     0.4  \\ 
    L10.C1  &   $ 384 \PLH 64 \PLH 1 \PLH 1 $    &  0.6  &     0.4  \\ 
    L10.C2  &   $ 384 \PLH 1 \PLH 3 \PLH 3 $     &  0    &     0    \\ 
    L10.C3  &   $ 96 \PLH 384 \PLH 1 \PLH 1 $    &  0.6  &     0.4  \\ 
    L10.SC  &   $ 96 \PLH 64 \PLH 1 \PLH 1 $     &  0.1  &     0.1  \\ 
    L11.C1  &   $ 576 \PLH 96 \PLH 1 \PLH 1 $    &  0.5  &     0.4  \\ 
    L11.C2  &   $ 576 \PLH 1 \PLH 3 \PLH 3 $     &  0    &     0    \\ 
    L11.C3  &   $ 96 \PLH 576 \PLH 1 \PLH 1 $    &  0.6  &     0.4  \\ 
    L12.C1  &   $ 576 \PLH 96 \PLH 1 \PLH 1 $    &  0.6  &     0.4  \\ 
    L12.C2  &   $ 576 \PLH 1 \PLH 3 \PLH 3 $     &  0    &     0    \\ 
    L12.C3  &   $ 96 \PLH 576 \PLH 1 \PLH 1 $    &  0.6  &     0.4  \\ 
    L13.C1  &   $ 576 \PLH 96 \PLH 1 \PLH 1 $    &  0.4  &     0.4  \\ 
    L13.C2  &   $ 576 \PLH 1 \PLH 3 \PLH 3 $     &  0    &     0    \\ 
    L13.C3  &   $ 160 \PLH 576 \PLH 1 \PLH 1 $   &  0.6  &     0.4  \\ 
    L14.C1  &   $ 960 \PLH 160 \PLH 1 \PLH 1 $   &  0.6  &     0.4  \\ 
    L14.C2  &   $ 960 \PLH 1 \PLH 3 \PLH 3 $     &  0    &     0    \\ 
    L14.C3  &   $ 160 \PLH 960 \PLH 1 \PLH 1 $   &  0.8  &     0.4  \\ 
    L15.C1  &   $ 960 \PLH 160 \PLH 1 \PLH 1 $   &  0.8  &     0.4  \\ 
    L15.C2  &   $ 1280 \PLH 320 \PLH 1 \PLH 1$   &  0.8  &     0.4  \\ 
    LINEAR  &   $d_\mathrm{out}  \PLH 1280 $     & 0.0   &     0.0  \\
    \bottomrule
    \end{tabular}
\end{table}

\end{document}